%% file: PaperForReview.tex
\documentclass[10pt,twocolumn,letterpaper]{article}

\usepackage{iccv}
\usepackage{times}
\usepackage{epsfig}
\usepackage{graphicx}
\usepackage{amsmath}
\usepackage{amssymb}
\usepackage{setspace}
\usepackage{multirow}
\usepackage{ctable} % for \specialrule command

\usepackage[pagebackref=true,breaklinks=true,letterpaper=true,colorlinks,bookmarks=false]{hyperref}

\iccvfinalcopy % *** Uncomment this line for the final submission

\def\iccvPaperID{6792} % *** Enter the ICCV Paper ID here

\ificcvfinal\fi

\begin{document}

\title{High-Fidelity Virtual Try-on with Large-Scale Unpaired Learning}

\author{
Han Yang\textsuperscript{1,2}\,\,\,\,\,\,\,\,\,\, 
Yanlong Zang\textsuperscript{3}\,\,\,\,\,\,\,\,\,\,
Ziwei Liu\textsuperscript{4} \,\,\,\,\,\,\,\,\,\,
\\
\centerline{
\textsuperscript{1}ETH Zurich\,\,\,\,\,
\textsuperscript{2}ZMO AI Inc.}\\
\centerline{
\textsuperscript{3}Zhejiang University\,\,\,\,\,
\textsuperscript{4}S-Lab, Nanyang Technological University }
\\
{\tt\small hanyang@ethz.ch, yanlongzang@zju.edu.cn, ziwei.liu@ntu.edu.sg}
}

\begin{figure}[htb]

\twocolumn[{
\renewcommand\twocolumn[1][]{#1}%
\maketitle
\vspace{-32pt}

\begin{center}
  \centering
  \includegraphics[width=1\textwidth]{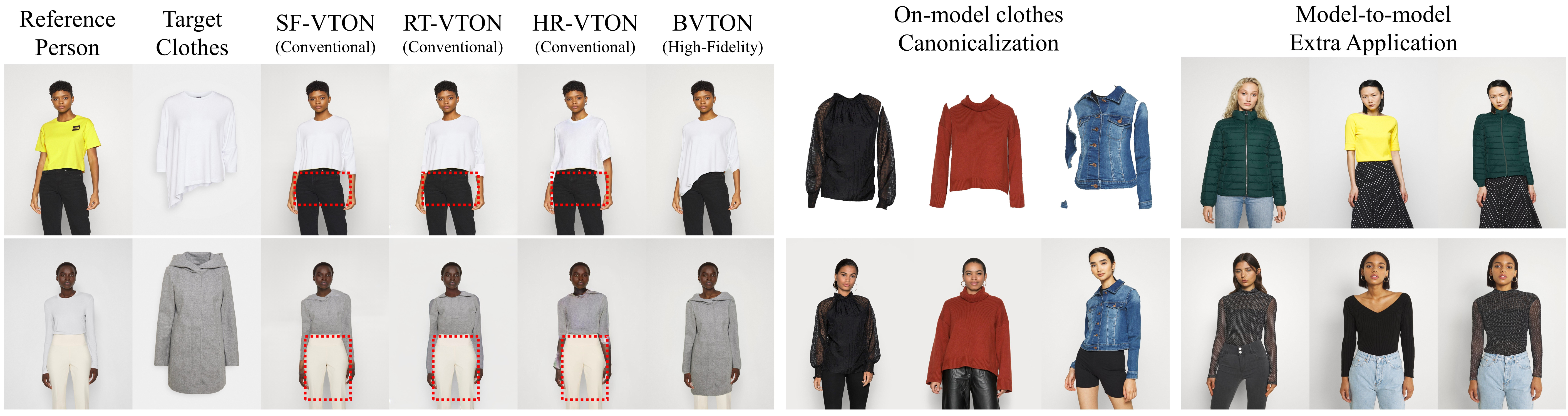}
  \vspace{-20pt}

\end{center}

\caption{Visual results showing the superiority of our high-fidelity try-on setting, boosted by large-scale unpaired learning. Our high-fidelity try-on pipeline, namely, BVTON preserves the full clothing details (clothing fidelity) including the asymmetric clothing bottom shapes. Conventional virtual try-on methods inherently fail to preserve the complete traits of the target clothes. Here we term \textbf{``conventional''} as the try-on setting used in previous try-on methods
% ~\cite{ACGPN,VITON_HD,SF-VTON,RT-VTON,HR-VTON,PF-AFN,DCTON}
that directly preserves the bottom clothes of the reference person regardless of the target clothing shapes. 
% Tuck-out effect can be trivially achieved by removing the bottom preservation area, but hence suffers from the ambiguity of paired training data. By learning a canonical proxy from the on-model-clothes, our method can utilize collecting friendly unpaired data to boost the accuracy of semantic prediction and naturally constructs an ambiguity-free scenario.
Besides, our method is also capable of an extra application for model-to-model try-on.}
}]
\label{fig:teaser}
\end{figure}
\ificcvfinal\thispagestyle{empty}\fi

%%%%%%%%% ABSTRACT
\begin{abstract}
% Image based virtual try-on targets at transferring an in-shop clothing image to a random person. Conventional pipelines require abundant ghost mannequin images paired with models wearing the same clothes. However, limited paired data largely hinders the development of learning-based virtual try-on methods.

Virtual try-on (VTON) transfers a target clothing image to a reference person, where clothing fidelity is a key requirement for downstream e-commerce applications.
However, existing VTON methods still fall short in high-fidelity try-on due to the conflict between the high diversity of dressing styles (\eg clothes occluded by pants or distorted by posture) and the limited paired data for training.
In this work, we propose a novel framework \textbf{Boosted Virtual Try-on (BVTON)} to leverage the large-scale unpaired learning for high-fidelity try-on.
Our key insight is that pseudo try-on pairs can be reliably constructed from vastly available fashion images. 
Specifically, \textbf{1)} we first propose a compositional canonicalizing flow that maps on-model clothes into pseudo in-shop clothes, dubbed canonical proxy. Each clothing part (sleeves, torso) is reversely deformed into an in-shop-like shape to compositionally construct the canonical proxy.
\textbf{2)} Next, we design a layered mask generation module that generates accurate semantic layout by training on canonical proxy. We replace the in-shop clothes used in conventional pipelines with the derived canonical proxy to boost the training process.
\textbf{3)} Finally, we propose an unpaired try-on synthesizer by constructing pseudo training pairs with randomly misaligned on-model clothes, where intricate skin texture and clothes boundaries can be generated.
Extensive experiments on high-resolution ($1024\times768$) datasets demonstrate the superiority of our approach over state-of-the-art methods both qualitatively and quantitatively. 
Notably, BVTON shows great generalizability and scalability to various dressing styles and data sources.

\end{abstract}
\input{sections/Introduction}

\input{sections/RelatedWork}

% \section{Related Works}

\input{sections/Method}

% \usepackage{ulem}

\input{sections/Experiments}

\input{sections/Conclusion}
%%%%%%%%% REFERENCES
{\small
\bibliographystyle{ieee_fullname}
\bibliography{egbib}
}
\newpage
\appendix

\input{appendix}

\end{document}

%% file: sections/Introduction.tex
\section{Introduction}

Virtual try-on, fitting the target clothes onto a reference person, have achieved great progress in recent years~\cite{ACGPN,SF-VTON,RT-VTON,HR-VTON}. Well-designed architectures are frequently proposed to build the relationship between the in-shop clothes and the reference person. However, we still observe three major problems in current try-on pipelines: 
\textbf{1)} \textbf{Low clothing fidelity} of current (\textbf{``conventional''}) try-on pipelines which ignore the actual shape of target clothing images. Clothing fidelity requires that the original clothing characteristics, especially the irregular designs, are well preserved.
\textbf{2)} \textbf{Insufficient data} on the widely-used clothes-model pairs. \textbf{3)} \textbf{Low-resolution} of the results generated by most of the current methods~\cite{ACGPN,DCTON,PF-AFN,RT-VTON,SF-VTON}. As resolution is one of the most important factors in downstream e-commerce applications, designing high-fidelity try-on pipelines with high-resolution output is essential.

% However, due to scarcity of the paired data, the model images with the corresponding in-shop clothes, the development of learning-based virtual try-on methods is coming to a bottleneck. 
% Moreover, high-resolution virtual try-on ($1024\times768$) is also rarely investigated in recent works~\cite{CPVTON,CPVTON+,ACGPN,DCTON,PF-AFN,SF-VTON}. For real application of virtual try-on, high-resolution is a basic requirement in order to leverage this task from non-profit research demo to commercial usage. Slightest artifacts and visual blurry,even unnoticeable in low-resolution, can be extremely obvious in high-resolution setting, posing great challenge in designing profitable pipelines for real commercial usage.

Earlier efforts in improving clothing fidelity lie on two key modules: characteristics-preserving deformation module and accurate semantic prediction module. From the very first Thin-plate Spline (TPS) based methods~\cite{VITON,CPVTON,CPVTON+,ACGPN}, various pioneering deformation methods are proposed such as Moving Least Squares (MLS) based method~\cite{RT-VTON} and flow-based methods~\cite{PF-AFN,SF-VTON}. With the maximized flexibility, flow-based methods can model any transformation with the regularized training objectives.
On the other hand, modeling the accurate after-try-on semantics is also crucial, as addressed in \cite{ACGPN,RT-VTON}. With limited paired data, learning the target semantic layout conditioned on the target clothes is a challenging problem. To alleviate the problem, RT-VTON proposes a tri-level attention mechanism by modeling the semantic prediction as a long-range correspondence learning that achieves state-of-the-art semantic accuracy in previous methods.
However, the aforementioned methods adopt the same conventional design which directly preserves the bottom clothes when fitting the target clothes at the reference person, ignoring the intricate shape details as well as the correct length of the target clothing, as in Fig.~\ref{fig:teaser}. 
A trivial fix can be achieved by removing the bottom clothes in the semantic prediction modules, but it severely suffers from another problem with the ambiguous wearing styles of the model images, which degenerates the stability of semantic prediction.

To tackle the aforementioned three major problems in current pipelines, we propose a novel framework, \textbf{Boosted Virtual Try-on (BVTON)} which generates high-resolution ($1024 \times 768$) results by leveraging the large-scale unpaired learning
for high-fidelity try-on. Specifically, BVTON consists of four major modules as shown in Fig.~\ref{fig:pipeline}. The first module, as the key of our solution, is the Clothes Canonicalization Module (CCM). The CCM predicts a compositional
canonicalizing flow that maps on-model clothes into pseudo in-shop clothes, dubbed canonical proxy. The second part is the Layered Mask Generation Module (L-MGM), which predicts the Layered semantic masks of the reference person wearing the target clothes. As opposed to prior arts, our L-MGM is not trained with in-shop clothes pairs but solely on large-scale fashion images, which is achieved by using the canonical proxy generated by the pre-trained CCM. Notably, our L-MGM is a plug-and-play module that can take the target clothes as input in the inference phase, so it can be used in any semantic-based pipeline~\cite{VITON_HD,ACGPN,RT-VTON}. The third part is the Mask-guided Clothes Deformation Module (M-CDM) that predicts deformation flow to warp the target clothes onto the reference person, guided by the layered semantic masks generated by L-MGM.
% Unlike flow-based methods which are vulnerable to the slight misalignment of the appearance flow, our flow-based canonicalization module is robust to misalignment and a roughly predicted flow is enough for further semantic layout learning with the canonical proxy. We can see from Fig.~\ref{fig:teaser} that the sleeves are not accurately aligned with the torso; the same misalignment can be devastating for the flow-based methods~\cite{PF-AFN,SF-VTON} without explicit semantic modeling.

% \begin{figure}[t]
% \begin{center}
% %\fbox{\rule{0pt}{2in} \rule{0.9\linewidth}{0pt}}
% \includegraphics[width=0.85\linewidth]{Images/BVTON_ambiguity.jpg}
% \vspace{-10pt}
% \end{center}
%    \caption{An example showing the ambiguous wearing style. 
%    % In the occluded style (right of the blue line), due to the variance of the model dressing style, the clothes lengths and the shape details of the target clothes are inconsistent with those of the on-model clothes, which causes inconsistent correspondence during the training. The occluded parts are red-boxed.
%    }
% \label{fig:ambiguity}
% \vspace{-10pt}
% \end{figure}

Finally, with the predicted segmentation, we propose an Unpaired Try-on Synthesizer Module (UTOM) by constructing pseudo training pairs with randomly misaligned on-model clothes, where intricate skin
texture and clothes boundaries can be generated. The UTOM also acts as a plug-and-play module that takes the warped clothes as input in the inference phase; it generates the final results according to the predicted layered masks. In this way, the spatial misalignment of the deformed clothes can be tolerated with guidance of layered semantic masks. 

% that  only needs unpaired data to train the fusion module in the conventional split-transform-merge setting~\cite{ACGPN,VITON_HD,RT-VTON}. Highly realistic skin details and coherent clothes textures can be achieved with UTOM boosted by the abundant unpaired data.

% In this work, we propose a novel framework Boosted Try-on (BVTON) which utilizes collecting-friendly unpaired data to better depict the semantic layout after the try-on process. The \textbf{key insight} is to learn a canonicalizing flow with limited paired data to map the on-model-clothes into the ghost-mannequin-like shapes, and thus the semantic predictor can be trained with the canonical proxy as input. Our design has two principled benefits: \textbf{1)} Large amount of unpaired data greatly boosts the semantic mining process. Unprecedented semantic accuracy is possible for the first time. \textbf{2)} Ambiguity-free learning is achieved with the canonical proxy, while the conventional paired setting suffers from the random wearing styles (\eg tuck-in, tuck-out) of models. Therefore, coherent tuck-out try-on effects are achieved for the first time. Ex

Our contributions can be summarized as follows: \textbf{1)} We design a principled try-on paradigm, \ie, BVTON, which generates high-resolution (1024 × 768) results by leveraging additional large-scale unpaired learning for high-fidelity try-on. Intricate clothing details such as laces, over-long clothes, and asymmetrical clothes bottoms can be well preserved. \textbf{2)} Our unified framework is the first cloth-to-model try-on approach that can adapt seamlessly to model-to-model virtual try-on without retraining. We demonstrate the incapability of baseline methods on model-to-model try-on setting in supp.. Note that we do not claim superiority over other model-to-model works, but only demonstrate as an extra application. \textbf{3)} We propose a novel unpaired try-on synthesizer that decouples the conventional try-on synthesis training from the limited paired data to boost the generative capabilities with large-scale unpaired learning on fashion images. \textbf{4)} BVTON greatly outperforms three lastest state-of-the-art methods~\cite{RT-VTON,SF-VTON,HR-VTON} across three different test sets (TEST1, TEST2 and VITON~\cite{VITON}). In the conventional setting (retaining bottom clothes), significant gains are achieved by 35.2\% in FID, 34.2\% in LPIPS, and 5.7\% in SSIM (TEST1 compared to \cite{HR-VTON}). Some extra baselines~\cite{PF-AFN,DAFlow} are given in quantitative results for reference.

% , and 7.3\% in SSIM (TEST2 compared to \cite{HR-VTON})

% and 34.2\% in LPIPS (TEST1 compared to \cite{HR-VTON}).

% FID: 35.2% in ZHUMAN,   34.0% in VITON,   24.1
% SSIM: 7.3% in ZHUMAN*, 6.1% in ZHUMAN

%% file: sections/RelatedWork.tex
\section{Related Works}

%For example, Physical Simulation~\cite{Physical Simulation} is an advanced synthesis performance with dynamic details, there are physical simulation works based on 3D information, which makes fashion synthesis more vivid. 

% \noindent \textbf{Fashion Image Synthesis.}\quad 
% %
% Fashion synthesis is an emerging research topic that attracts increasing attention due to its close relationship with the industry. From fashion style transfer~\cite{Fashion_Style_Generator,swapgan} to interactive fashion design~\cite{Interactive_Sketching_System_for_Fashion,fashion_editing,Edit_Like_A_Designer}, generative models are showing great potential in the fashion industry. Moreover, generating clothes with certain conditions is also a challenging problem due to the complex texture characteristics, as in \cite{Be_your_own_prada,Poly-GAN}.

\noindent \textbf{Image-based Virtual Try-on.}\quad
Image-based virtual try-on is focused on transferring the specified in-shop clothes to the reference person while preserving the clothing shapes and the reference person posture as well as identity. Due to the unaffordable cost of cloth simulating and physically based rending, performing highly-realistic virtual try-on within the 2D image scenario has been a hot research topic with its great commercial profit potential.

Earlier methods~\cite{VITON,CPVTON,CPVTON+} use coarse shapes, pose map, and TPS warping to perform image-based virtual try-on. Without explicit semantic layout representation, the clothes-skin boundaries are blurry and the final results are far from photo-realistic. ACGPN~\cite{ACGPN} later proposes a semantic layout representation that decouples the learning process of shape modeling and texture synthesis, achieving photo-realistic try-on results for the first time. However, ACGPN still suffers from predicting stable and accurate after-try-on semantics, ascribing to the unawareness of the long-range correspondence between the reference person and the target in-shop clothes. RT-VTON~\cite{RT-VTON} proposes a Tri-Level Transform block that successfully handles the non-standard clothing shapes, which is a huge step towards the robust try-on scheme. Purely flow-based methods such as SF-VTON~\cite{SF-VTON} directly predict the appearance flow to deform the target clothes and generate the try-on results accordingly without explicit semantic layout representation, which fail to generate realistic results in a high-resolution ($1024\times 768$) scenario; it also suffers from the spatial misalignment of warped clothes without guidance of semantic layout. VITON-HD~\cite{VITON_HD} and HR-VTON~\cite{HR-VTON} focus on high-resolution virtual try-on that produce the state-of-the-art highly-realistic results in high-resolution virtual try-on. However, the aforementioned methods strongly rely on the paired clothes-model data, which largely hinders the development of the data-driven approach to promote image-based virtual try-on.

%
% VITON~\cite{VITON} and deforms the in-shop clothes by Thin-plate spline (TPS) transformation, and generate the final output given the coarse shape of hum handle complex warping problems.
% SF-VTON~\cite{SF_VTON} is a flow-based models, using a novel global appearance flow estimation model to pay more attention to local garment and context separately. Despite great improvements parse-free based have been achieved, the artifacts in final generated images caused by ... are hard to be solved.
% ACGPN~\cite{ACGPN} proposed a new framework that first predicts the semantic layout of the reference image after try-on and then warp the ghost mannequin according to the target cloth region, which generating realistic-looking results but not alleviating the ambiguity between the garment and reference model.
% RT-VTON~\cite{RT_VTON} is a upgraded version of ACGPN, combining the local gated attention with global correspondence learning to gradually refine the tri-level feature codes, which predicts the accurate semantic layout for further generation.
% HR-VTON~\cite{HR_VITON} proposed a unified condition generator which allows semantics prediction and garment transformation simultaneously, which bridges the information gap between garment and reference model.
% However, due to pair-data paucity, parser-based model is difficult to generate accurate and sharp semantics and smooth skin, which is our main focus.

\begin{table}
\setlength{\baselineskip}{2em} % 行间距
\renewcommand\tabcolsep{1pt} %
\small
\begin{center}
\begin{spacing}{1}

\begin{tabular}{llcccccccc}
\specialrule{.15em}{.05em}{.05em} 
&&\cite{VITON}&\cite{ACGPN}&\cite{DCTON}&\cite{SF-VTON}&\cite{RT-VTON}&\cite{HR-VTON}&\cite{PASTA-GAN}&Ours\\
\specialrule{.15em}{.05em}{.05em} 
\parbox[t]{2mm}{\multirow{2}{*}{\rotatebox[origin=c]{90}{\tiny Setting}}}&Model-to-model& $\times$  &$\times$&$\times$&$\times$&$\times$&$\times$&$\surd$&$\surd$\\
&Cloth-to-model& $\surd$ &$\surd$&$\surd$&$\surd$&$\surd$&$\surd$&$\times$&$\surd$\\
% &Use Segmentation& $\times$ &$\times$&$\times$&$\surd$&$\surd$\\
\hline

\parbox[t]{2mm}{\multirow{3}{*}{\rotatebox[origin=c]{90}{\tiny Supervision}}}&Only Paired& $\surd$ &$\surd$&$\surd$&$\surd$&$\surd$&$\surd$&$\times$&$\times$\\
&Only Unpaired& $\times$ 
&$\times$&$\times$&$\times$&$\times$&$\times$&$\surd$&$\times$\\
&Paired {\footnotesize+ Unpaired boosting}& $\times$ &$\times$&$\times$&$\times$&$\times$&$\times$&$\times$&$\surd$\\
\hline

\parbox[t]{2mm}{\multirow{3}{*}{\rotatebox[origin=c]{90}{\tiny Contribution}}}
&Network Architecture& $\times$ &$\times$&$\times$&$\surd$&$\surd$&$\surd$&$\surd$&$\times$\\
% &Modular Design& $\surd$ &$\surd$&$\surd$&$\surd$&$\surd$&$\surd$&$\times$&$\times$\\
&Pipeline formulation& $\surd$ &$\surd$&$\surd$&$\times$&$\times$&$\times$&$\surd$&$\surd$\\
&Plug-and-play& $\times$ &$\times$&$\times$&$\times$&$\times$&$\times$&$\times$&$\surd$\\

\specialrule{.15em}{.05em}{.05em} 
\end{tabular}
\vspace{5pt}

\caption{Comparison of representative virtual try-on methods: VITON~\cite{VITON}, ACGPN~\cite{ACGPN},DCTON~\cite{DCTON},SF-VTON~\cite{SF-VTON},RT-VTON~\cite{RT-VTON},HR-VTON~\cite{HR-VTON} and PASTA-GAN~\cite{PASTA-GAN}.}
\label{tab:comparison}

\end{spacing}
\vspace{-25pt}

\end{center}
\end{table}

\noindent \textbf{Model-to-model Virtual Try-on.}\quad
Model-to-model try-on aims at transferring the clothes from the target model onto the reference person.
Swapnet~\cite{Swapnet} proposes an unpaired model-to-model try-on pipeline that applies random affine transformations to construct the training pairs.
M2E-Tryon~\cite{M2E_Tryon} extracts the UV texture from densepose~\cite{densepsoe} representation and generates the coarse output with the UV warped texture. Final try-on results are generated by combining the identity of the reference person and the warped clothing texture.
O-VITON~\cite{O_VITON} applies an auto-encoding training scheme to decouple the texture and the clothing shapes. An online optimization method is proposed to refine the clothing texture.
PASTA-GAN~\cite{PASTA-GAN} is the state-of-the-art model-to-model try-on pipeline which utilized patch representation for unpaired try-on. Due to the discrete representation of rule-based patch extraction, it is not applicable to the cloth-to-person try-on.

We show the differences of the influential works on try-on in Tab.~\ref{tab:comparison} for better understanding of the various design choices in this area.
%
% The biggest problem behind the model-to-model try-on method is the requirement for a model actually wearing the target clothes. Those methods cannot transfer the in-shop clothes onto the reference person. 

%% file: sections/Method.tex
\section{Boosted Virtual Try-on}

\begin{figure*}[t]
\begin{center}
\includegraphics[width=0.95\linewidth]{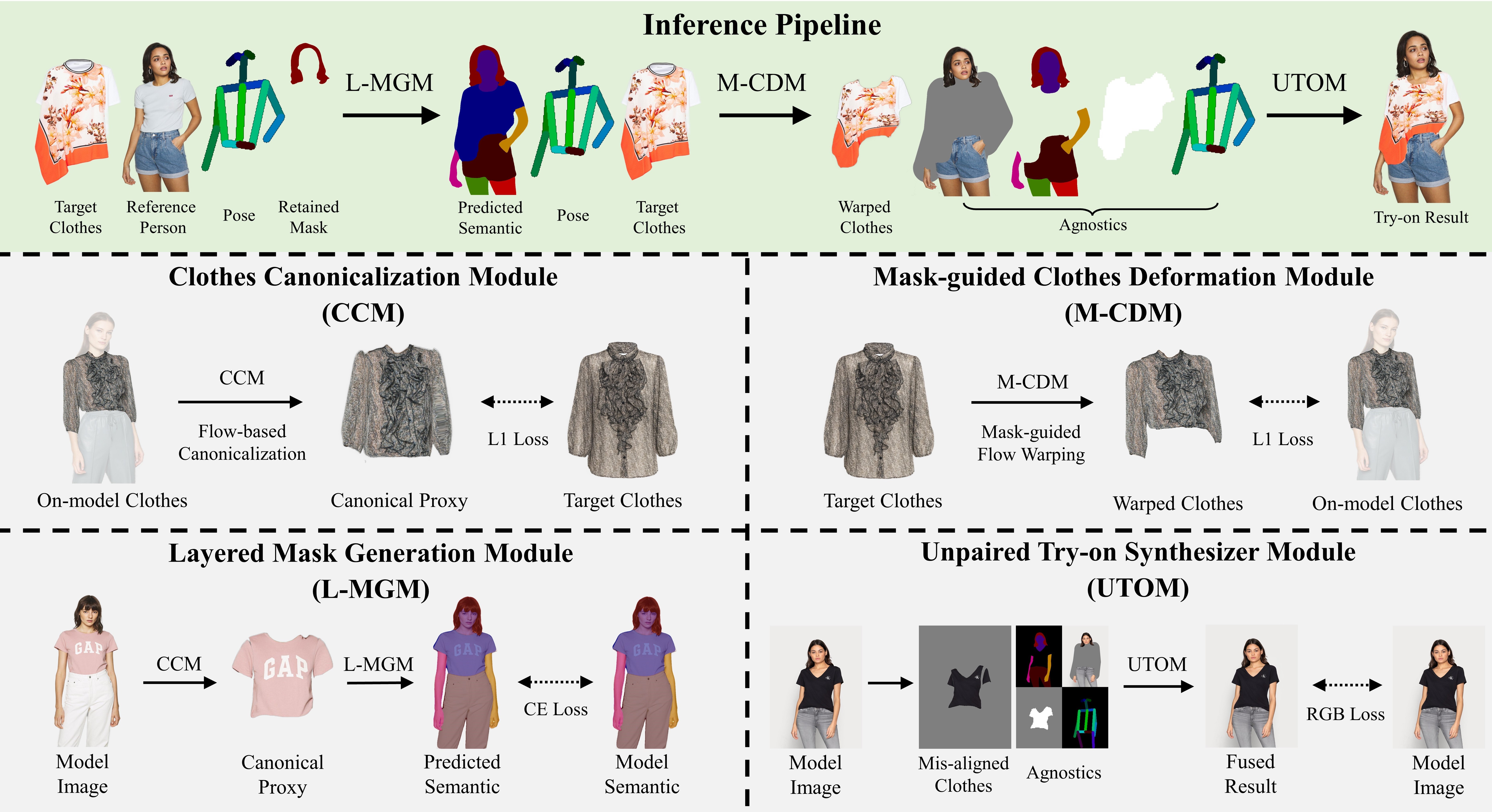}
\vspace{-10pt}
\end{center}
   \caption{The overall pipeline of BVTON including the training and inference workflows. Network details are given in Fig.~\ref{fig:networks}. CCM is first trained with paired data to predict the compositional canonicalizing flow for on-model clothes. We then extract the canonical proxies for the large-scale fashion images, and train the L-MGM with the proxies instead of the in-shop clothes. With predicted layered semantic masks, clothes can be warped accordingly in M-CDM. Finally, UTOM fuses the agnostics and the warped clothes to generate the try-on results.
%   Notably, M-CDM depends on clear mask guidance and UTOM strictly follows the predicted semantic, which explains why only limited paired data can suffice for the M-CDM training.    
   }
\label{fig:pipeline}
\vspace{-10pt}
\end{figure*}

\noindent \textbf{Framework Overview.}\quad 
To achieve high-fidelity try-on, BVTON follows the conventional semantic layout-based pipeline (Fig.~\ref{fig:pipeline}) as \cite{RT-VTON,HR-VTON}, which first predicts the semantic segmentation, dubbed layered masks of the after-try-on person with Layered Mask Generation Module (L-MGM) and predicts the deformation appearance flow with Mask-guided Clothes Deformation Module (M-CDM) to warp the target clothes, which are later fed into an Unpaired Try-on Synthesizer Module (UTOM) to generate the final output. The significant differences opposed to common practice are that our L-MGM and UTOM are trained by large-scale unpaired learning, where no clothes-model image pairs are required during the training. Paired data are only used in CCM and M-CDM to train the appearance flow.

% Notably, the data limitation of paired data is not the bottleneck for CCM and M-CDM for the following reasons: \textbf{1)} The canonicalizing flow is invulnerable to slight spatial misalignment, as shown in Fig.~\ref{fig:teaser}, where the sleeves are not accurately aligned with the torso part. \textbf{2)} The regularization term used in general flow learning naturally retains the original shape of clothes, which is the base of our ambiguity-free training, as in Fig,~\ref{fig:ambiguity}. \textbf{3)} M-CDM strictly follows the predicted semantic layout which is a relaxed learning objective compared to the conventional flow-based methods~\cite{PF-AFN,SF-VTON}, and our UTOM generates try-on results according to the semantic layout which makes BVTON invulnerable to inaccurate clothes deformation.
\subsection{Compositional Clothes Canonicalization}
Previous semantic layout-based methods~\cite{ACGPN,RT-VTON,VITON_HD,HR-VTON} demonstrate great superiority in generating clear clothes boundaries and realistic results, while the accuracy of the predicted layout becomes the main bottleneck for deploying the methods into real practice. Two major problems affecting semantic accuracy are first the limited access to paired data, and second, the occlusions caused by ambiguous wearing styles. In terms of conventional try-on~\cite{ACGPN,DCTON,VITON_HD}, bottom clothes are given as retained area so the aforementioned ambiguous wearing styles cause no trouble; however, when we need to exhibit all the intricate shape details such as asymmetrical clothes bottoms or overlong clothes that overlay the upper clothes above the  bottom clothes, modeling the random semantics with no prior clues tends to converge to averaged shapes.

To address the problems, we propose a Clothes Canonicalization Module (CCM) to predict a compositional canonicalizing flow that maps on-model clothes into pseudo in-shop clothes, dubbed canonical proxy, and the L-MGM can be thus trained with the pseudo pairs. 
% Without the straight forward ``T-Pose"" analogy as in the 3D body senario~\cite{SMPL,SMPLX,SCANImate}, we need to define the canonical version of the clothes. Interestingly, the clothing image (ghost mannequin) used in conventional try-on pipelines~\cite{VITON,CPVTON,ACGPN} is just a suitable candidate to be recognized as the canonical version of the clothes on the model.
% Unlike the previous flow-based try-on pipelines~\cite{PF-AFN,SF-VTON} that predict the appearance flow to warp the clothing image given the reference pose, we reverse the process to learn the canonicalizing flow to deform the on-model clothes into the in-shop clothes shape.
We have several benefits with this design: \textbf{1)} The accuracy of canonicalizing flow is not highly demanded, since the slight misalignment of spatial location will not directly affect the final try-on results, while flow-based methods like SF-VTON~\cite{SF-VTON} directly preserve most of the warped clothes without explicit semantic modeling. \textbf{2)} Due to regularization in flow learning, the overall shapes of the on-model clothes are fully preserved so that the target try-on semantics are deterministic given the canonical proxy as input in L-MGM training.

%% flow formula losses
To stabilize training, we first follow the reverse mapping scheme as proposed in \cite{clothformer} to map the occlusion area (bottom clothes and hair) onto the in-shop clothes with semi-rigid deformation~\cite{RT-VTON}. With the predicted semi-rigid deformation parameters $\theta_t$ and the reverse parameters $\theta^*_t$ (by swapping the target and source control points), we can remove the occluded area in the target clothes $C_t$ by: 
\begin{equation}
\hat{C}_t=C_t \odot (1-W(M_o,\theta^*_t)),
\end{equation}
where $W(\cdot,\cdot)$ denotes the back warping operation (also known as grid sampling) and $M_o$ denotes the occlusion area of hair and bottom clothes. $\odot$ denotes the element-wise multiplication. Then our canonicalizing flow estimation can be trained by reversing the conventional flow estimation objective to warp the on-model clothes $C_m$ onto the occluded target clothes $\hat{C}_t$, as:
\begin{equation}
\mathcal{L}_{cano}^{flow}=\lVert \hat{C}_t- W(C_m,f_{cano})  \rVert_1,
\label{equa:cano}
\end{equation}
where the canonicalizing flow $f_{cano}$ is conditioned on the on-model clothes and the human pose. Style-based flow estimators are adopted following~\cite{SF-VTON}, with modulated convolution and gradually refined flow estimation pipeline.

With the canonicalizing flow estimator, we can derive the canonical proxies from the large-scale fashion images, which are in regular in-shop clothing shape while retaining the slightest shape variance credited to the regularizing terms in flow learning. The derived canonical proxies are later used to train the most essential part of semantic-based models, the Layered Mask Generation Module (L-MGM).
\subsection{Layered Mask Generation Module}
The semantic layout representation, as first proposed in \cite{ACGPN,sievenet}, greatly improves the try-on quality from earlier blurry results~\cite{CPVTON,VITON} to photo-realistic images, which is now a conventional design for image-based virtual try-on.  Although semantic-based methods excel at generating clear clothes-skin boundaries, they suffer from the accuracy of the predicted layout. To address the problem, RT-VTON proposes a Tri-Level Transform block that integrates gated convolution with long-range attention modeling to bridge the gap between the target clothes and the reference person. Surprising results are achieved by Tri-Level Transform; however, RT-VTON only shows the conventional try-on setting with retaining the bottom clothes of the reference person, which blocks itself from performing high-fidelity try-on for broader applications. Moreover, the limited access to paired data also hinders the accuracy of predicted semantic layout, especially for the non-standard cases with puff sleeves, lace decorations, or intricate collar shapes.

To address the aforementioned problem, our L-MGM can be trained with large-scale fashion images with the canonicalized on-model clothes as input to predict the semantic layout in an auto-encoding manner, as shown in Fig.~\ref{fig:pipeline}. We apply the Tri-Level blocks following \cite{RT-VTON} to build our L-MGM. As shown in Fig.~\ref{fig:pipeline} (c), L-MGM takes the canonical proxy $W(C_m,f_{cano})$, the human pose $P$ and retained masks $M_\omega$ as input to predict the layered masks (semantic layout).
%% COPY RTVTON formula
The feature codes are first extracted from three individual encoders, and the Tri-Level blocks update the feature codes with gated masks as well as non-local attention modeling.
Staring from the $0^{th}$ codes, the $t^{th}$ clothes code $F_t^C$, the pose code $F_t^P$ and parsing code $F_t^S$ are updated as follows: We first compute the local gating masks from the $(t-1)^{th}$ parsing code $F_{t-1}^S$ by
\begin{equation}
\begin{split}
&M_t^C=\sigma(\text{conv}(F_{t-1}^S)),\\
&M_t^P=\sigma(\text{conv}(F_{t-1}^S)),
\end{split}
\label{equa:attention}
\end{equation}
where $\sigma$ indicates the element-wise $sigmoid$ function and $\text{conv}$ indicates the convolutional layers. Then we compute the correlation matrix $\mathcal{M}_C^t$ used in the long-range correspondence modeling. The feature codes are flattened to get $x'_t{}^P \in \mathbb{R}^{HW\times C}$ (pose), and $x'_t{}^C \in \mathbb{R}^{HW\times C}$ (clothes).
\begin{equation}
    \mathcal{M}_C^t(u,v)=\frac{\hat{x}_t^C(u)^T\hat{x}_t^P(v)}{ \left\lVert \hat{x}_t^C(u)\right\rVert_2 \left\lVert \hat{x}_t^P(v) \right\rVert_2},
\end{equation} 
 where $\hat{x}_t^C(u)$ and $\hat{x}_t^P(v)$ indicate the channel-wise centralized features. Then we transform the flattened code $x_t^C$ by \begin{equation}
    \Bar{x}_t^C=softmax_{v}(\alpha \mathcal{M}_C^t) x_t^C,
\label{equa:corr_trans}
\end{equation}
where $\alpha$ is a re-weighting term in \cite{RT-VTON}. We reshape $\Bar{x}_t^C$ to get $ \Bar{F}_t^C$. With the transformed code and the gating masks, we update the codes as follows:
\begin{equation}
    \begin{split}
    &F_t^C=M_t^C \odot \text{conv}(F_{t-1}^C)+F_{t-1}^C\\
    &F_t^P=M_t^P \odot \text{conv}(F_{t-1}^P)+F_{t-1}^P,\\
        &F_t^S=\gamma(\Bar{F}_t^C) \odot F_{t-1}^S + \beta (\Bar{F}_t^C),
    \end{split}
\end{equation}
where $\gamma(\cdot)$ and $\beta(\cdot)$ indicate the modulations in Spatial Feature Transform (SFT)\cite{SFT}. $\odot$ denotes element-wise multiplication. The full training objective for our L-MGM is as follows:
\begin{equation}
    \mathcal{L}_{layout}=\lambda_{1}\mathcal{L}_{CE}+\lambda_{2} \mathcal{L}_{cGAN}+\lambda_{3}\mathcal{L}_{TV}+\lambda_{4}\mathcal{L}_{warp},
\end{equation}
where $\mathcal{L}_{CE}$ denotes the per-pixel cross entropy loss, $\mathcal{L}_{cGAN}$ denotes the conditional GAN loss, and $\mathcal{L}_{TV}$ indicates the total variance term. $\mathcal{L}_{warp}$ is the attention warping loss used in \cite{RT-VTON}, parameterized in $\mathbb{L}_1$ distance.  The Gumbel softmax trick~\cite{gumbel} is applied to discretize the output for differentiable training in computing $\mathcal{L}_{cGAN}$.

For the high-fidelity setting, we randomly give the bottom clothes of the reference person at probability $p$. With this design, L-MGM, in the inference phase, first predict the  target clothing mask $M_c^1$ with the target clothes $C_t$
and we get the occluded bottom clothes mask $M_b^o=M_b\odot( 1-M_c^1),$ where $M_b$ is the bottom clothing mask of the reference person. $M_b^o$ is then added to the retain masks $M_\omega$. Then we inference the L-MGM for another time, to generate the layered masks according the correctly occluded bottom mask for high-fidelity.

% Tuck-out setting is simply achieved by randomly removing the bottom clothes in the input retained area. With the canonical proxy, we can break the data bottlenecks of paired data to improve the accuracy of semantic prediction. As shown in Fig.~\ref{fig:ambiguity}, canonical proxy retains the model-specific shape variations due to the flow regularization, which results in an ambiguity-free training scenario in our large-scale unpaired training for high-fidelity try-on. 

\begin{figure}[t]
\begin{center}
\includegraphics[width=1\linewidth]{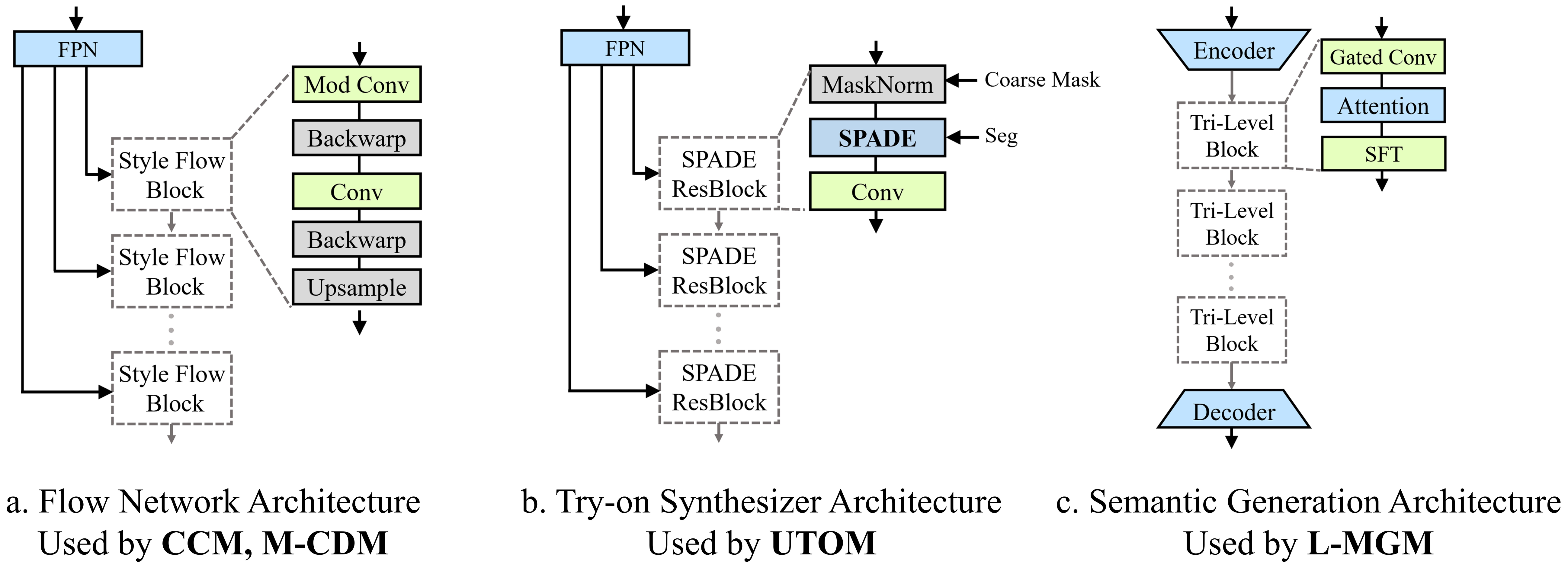}
\vspace{-10pt}
\end{center}
   \caption{The network design details of the modules used in BVTON. FPN denotes the feature pyramid network.}
\label{fig:networks}
\vspace{-10pt}
\end{figure}

\subsection{Mask-guided Clothes Deformation}
With the predicted semantic layout, our M-CDM module predicts the deformation flow to warp the clothes according to the layout. This is a classic design as in semantic-based methods~\cite{ACGPN,RT-VTON}; the only difference is that we use the appearance flow instead of TPS or semi-rigid deformation to better fill the predicted clothes mask. M-CDM is trained with paired data, which actually suffices with mask guidance and the spatial misalignment will not affect the final synthesis of UTOM guided by layered semantic masks. On the other hand, SF-VTON directly preserves the warped clothes, which can be vulnerable to flow inaccuracy. As shown in Fig.~\ref{fig:other_data} (row.~1), the sleeves of the target clothes are not correctly aligned with the reference person in the result of SF-VTON.

Concretely, the training objective of M-CDM is similar to CCM, as:
\begin{equation}
\mathcal{L}_{deform}^{flow}=\lVert C_m- W(\hat{C}_t,f_{deform})  \rVert_1,
\label{equa:deform}
\end{equation}
where the flow $f_{deform}$ is conditioned on the occluded target clothes $\hat{C}_t$, the target clothing masks and the pose.

\begin{figure*}[t]
\begin{center}
\includegraphics[width=1\linewidth]{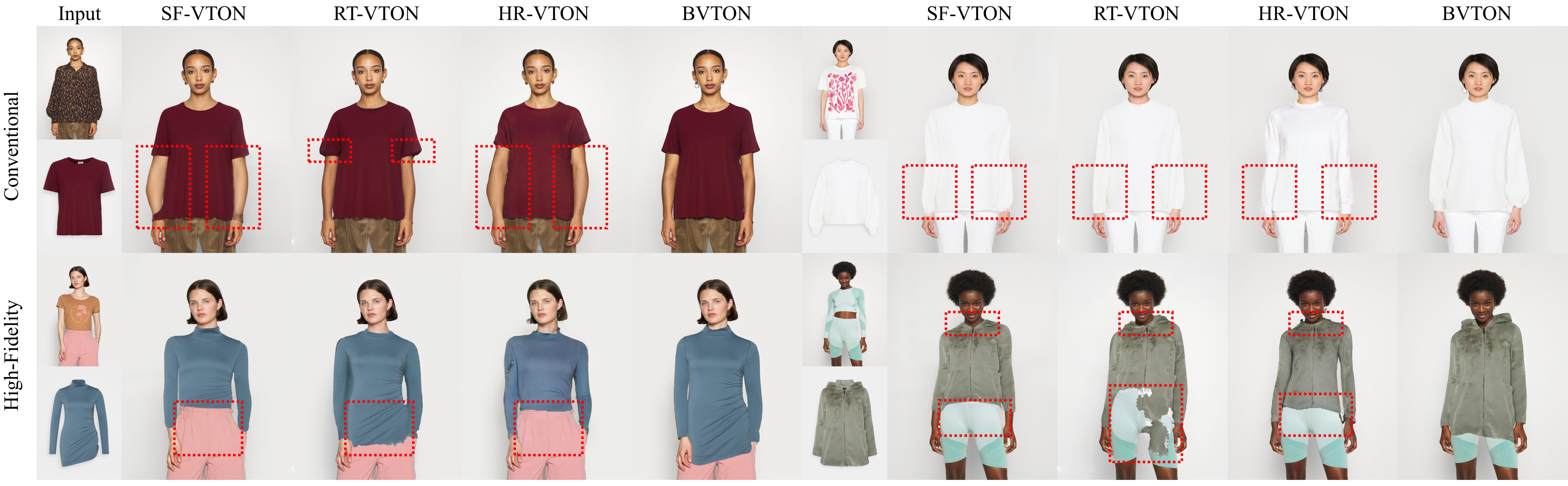}
\vspace{-20pt}
\end{center}
   \caption{Visual comparison of four virtual try-on methods. The first row shows the conventional setting that directly preserves the bottom clothes, and the second row shows the high-fidelity try-on results. With the help of vastly available fashion images, BVTON can generate realistic results with high clothing fidelity and remarkable skin details. Especially, BVTON generates realistic skin-clothes boundaries instead of simply overlaying the clothes onto the reference person. Visual artifacts are red-boxed.}
\label{fig:Comparison}
\vspace{-10pt}
\end{figure*}

\subsection{Unpaired Try-on Synthesizer}

The conventional split-transform-merge scheme~\cite{ACGPN,RT-VTON,VITON_HD,HR-VTON} consists of a Try-on Synthesizer Module (TOM) to fuse the warped clothes and the preserved body parts, guided by the predicted semantic layout. The common practice to train such a TOM is generally the same as the inference phase, which requires paired data to generate coherent synthesized results. 

Our proposed Unpaired Try-on Synthesizer Module (UTOM), on the other hand, adopts a novel large-scale unpaired learning to generate high-fidelity try-on results. We apply the random affine transformation onto the on-model clothes to construct pseudo training pairs. UTOM can thus generate realistic results boosted by vastly available fashion images.
% Furthermore, we observe that the predicted semantic layout is not as sharp as the ground-truth segmentation, so we downsample the clothes mask to low-resolution to generate sharp clothes boundaries in inference phase.

Specifically, we define the random affine augmentation operation as $A(I,r)$, which samples a value $v\in [-r,r]$ and transforms the point $p_I$ from image $I$ to $p'_I$ by:
\begin{equation}
    p'_I=R(\frac{v\pi}{180})p_I+v,
\end{equation}
where $R(\cdot)$ denotes the 2D rotation matrix. And thus we can derive the misaligned clothes $C_m^{mis}$ from the model image as the input of UTOM training as:
\begin{equation}
    C_m^{mis}=M_c \odot A(C_m,\alpha_{aug}) \odot A(M_c,\beta_{aug}), 
\end{equation}
where $\alpha_{aug}$ are $\beta_{aug}$ random affine parameters and $\beta_{aug}>\alpha_{aug}$ to create more misalignment to enhance the inpainting ability of the network. Empirically, $\alpha_{aug}$ is set as 1 and $\beta_{aug}$ is set as 4. $M_c$ denotes the mask of the on-model clothes $C_m$. We also observe the predicted masks from L-MGM are always not as sharp as the ground-truth masks, which can be also observed from the results from RT-VTON as in Fig.~\ref{fig:Comparison} (row.~1). To reduce the strong dependency on the predicted masks on the boundaries, we apply a degeneration method on $C_m \in \mathbb{R}^{H\times W \times 3}$ with resizing operator $rs(\cdot)$ with area interpolation and binarization operator $B(\cdot)$ as
\begin{equation}
C_m'=B\left(rs\left(rs\left(C_m,\left(H_\alpha,W_\alpha\right)\right),(H,W)\right)\right),
\end{equation}
where $H_\alpha$ and $W_\alpha$ are set as relative small numbers, and $C_m'$ is the degenerated mask. Empirically, we set $H_\alpha$ to be 100 and $W_\alpha=\frac{3}{4}H_\alpha$ for ratio concerns. We follow the design as used in \cite{VITON_HD,HR-VTON} to build the architecture with SPADE\cite{SPADE} blocks. We replace the clothing mask $C_m$ in semantic layout $S$ by $C_m'$, constructing $S'$ which is the input to obtain the SPADE modulation parameters $\gamma$ and $\beta$. The output activation $h'^i$ at site ($n\in N, c\in C^i,y\in H^i, x\in W^i$) is computed by:
\begin{equation}h'^i_{n,c,y,x}=\gamma_{c,y,x}^i(S')\frac{h^i_{n,c,y,x}-\mu_{n,c}^{i,k}}{\sigma_{n,c}^{i,k}}+\beta_{c,y,x}^i(S'),
\end{equation}
where $\mu_{n,c}^{i,k}$ and $\sigma_{n,c}^{i,k}$ are the mean and standard deviation of the activation $h^i$ in channel $c$. $k$ indicates whether the site is in the degenerated mask $C'_m$ or not, following the MaskNorm~\cite{VITON_HD}, a vairant of InstanceNorm~\cite{Instancenorm} by computing heterogeneous statistics according to the given masks. Instead of directly feeding $C'_m$ to the input, we integrate $C'_m$ in modulation calculation and normalization to reduce the dependency on the clothing mask in training and leave freedom for the network to adapt to the data for generating results with clear clothes boundaries, as in Fig.~\ref{fig:Comparison} (row.~1). The full objective of UTOM is
\begin{equation}
    \mathcal{L}_{rgb}=\lambda_{5}\mathcal{L}_{vgg}+\lambda_{6} \mathcal{L}_{L1}+\lambda_{7} \mathcal{L}_{cGAN}+\lambda_{8}\mathcal{L}_{feat},
\end{equation}
where $\mathcal{L}_{vgg}$ denotes the perceptual loss~\cite{vgg}, and $\mathcal{L}_{feat}$ denotes the feature matching loss~\cite{pix2pixhd}.

% Semantic-based methods may suffer from the smoothness of predicted semantic layout, as RT-VTON fails to generate sharp clothes boundaries

% With the misaligned clothes and the person agnostic representation, as shown in Fig.~\ref{fig:pipeline}, UTOM synthesize the final try-on output.
% %show the random affine equations

%% file: sections/Experiments.tex
\section{Experiments}

\begin{figure}[htb]
\begin{center}
\includegraphics[width=0.85\linewidth]{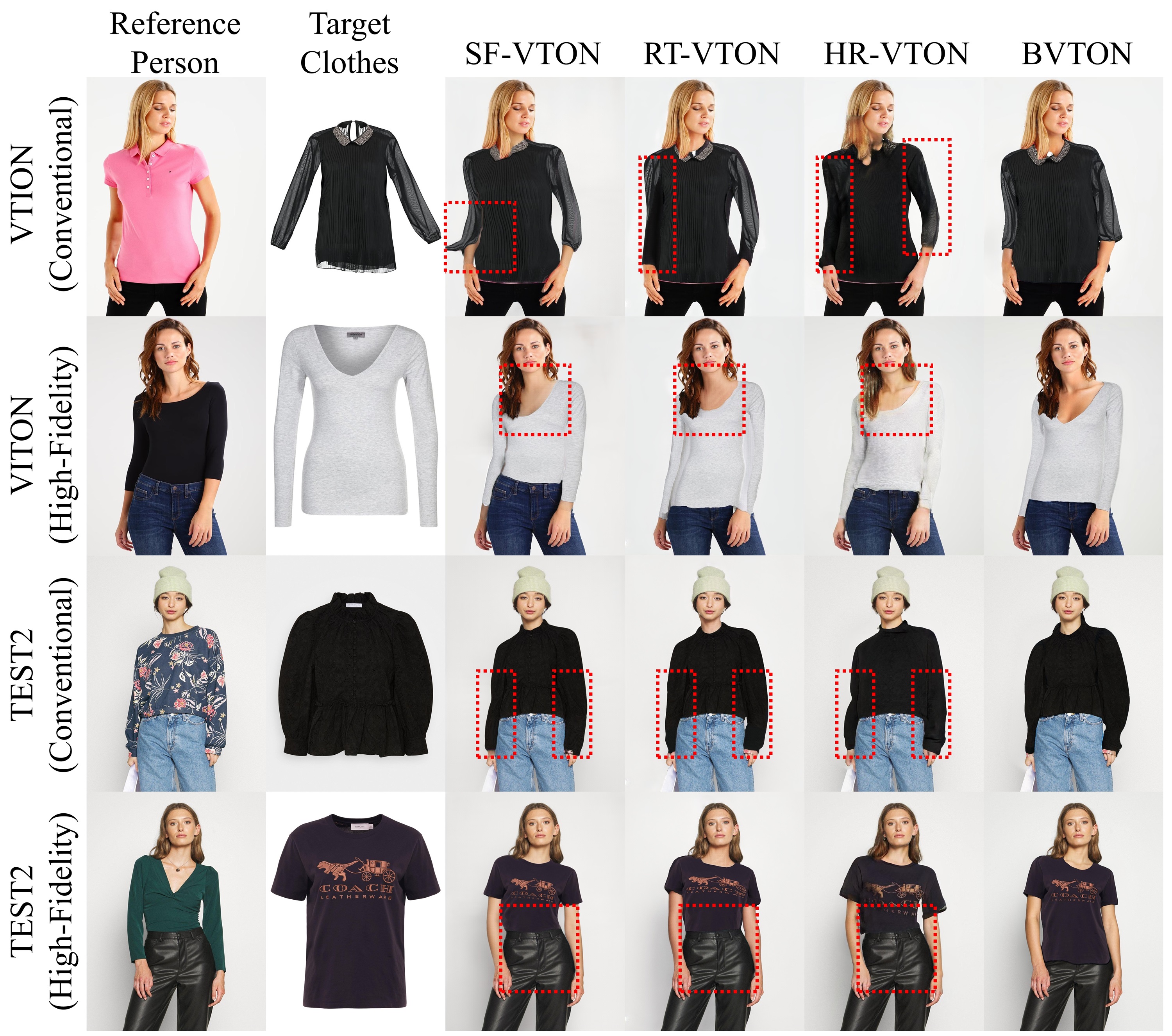}
\vspace{-10pt}
\end{center}
   \caption{Visual comparison of four virtual try-on methods in VITON and TEST2 test set. We can see that BVTON generalizes well on out-of-domain test data in both the conventional and the high-fidelity setting, which demonstrates the generalizability and scalability of our method. Visual artifacts are red-boxed.}
\label{fig:other_data}
\vspace{-10pt}
\end{figure}

\subsection{Experimental Setup}
\noindent\textbf{Datasets.}\quad
We collect a high-resolution ($1024\times 768$) upper-body and front-view fashion image dataset from the internet with 18,327 paired data, \ie, PAIRED Dataset.  The pairs are split into a training set and a test set with 15,527 and 2,800, respectively. We call the test set of PAIRED as TEST1 set. Besides, we collect 50,415 vastly available upper-body fashion images without corresponding in-shop clothes.
To further evaluate the generalization ability, we directly run inference on two more datasets trained on PAIRED: \textbf{1)} the High-resolution ($1024 \times 768$) VITON~\cite{VITON} dataset, the high-resolution version of the conventional benchmark, \textbf{2)} the TEST2 test set of 2800 data, which are automatically aligned upper-body images from full-body images collected from Internet. We assume different distributions between TEST1 and TEST2 since the data from TEST2 are originally full-body images.

\noindent \textbf{High-Fidelity Try-on.}\quad To evaluate the effectiveness of our high-fidelity setting, we select 50 models from each of the three datasets with completely visible bottom clothes to apply for target clothes with various lengths.

\noindent \textbf{Experimental Details.}\quad The batch-sizes for CCM, L-MGM, M-CDM and UTOM are set to 4, 4, 4 and 2, respectively. All modules above are trained for 20 epochs and adopted the Adam optimizer with $\beta_1 = 0.5$ and $\beta_2 = 0.999$. In CCM and L-MGM, the learning rates of semi-rigid deformation~\cite{RT-VTON} are 0.01 while the learning rate of flow estimator is set to 1e-6 in CCM and 5e-5 in M-CDM. The target clothing images are divided into three compositional parts (sleeves and torso) during training with off-the-shelf parser. Mask warping loss is applied the same way in CDM and CCM as Equa.~\ref{equa:cano} and Equa.~\ref{equa:deform}, and we omit it in expressions for simplicity. The back collars of target clothes are removed in advance for all methods. In L-MGM, the learning rates of the generator and the discriminator are set to 1e-4 and those of the UTOM are set to 1e-4 and 4e-4, respectively. All the codes are implemented in PyTorch and trained on 1 Tesla A40 GPU. Distillation trick is not applied for all methods for fair comparison. The loss terms are 10, 1, 0.1, 1, 10, 1, 1, and 1, from $\lambda_1$ to $\lambda_8$, respectively.

\begin{figure}[t]
\begin{center}
\includegraphics[width=0.95\linewidth]{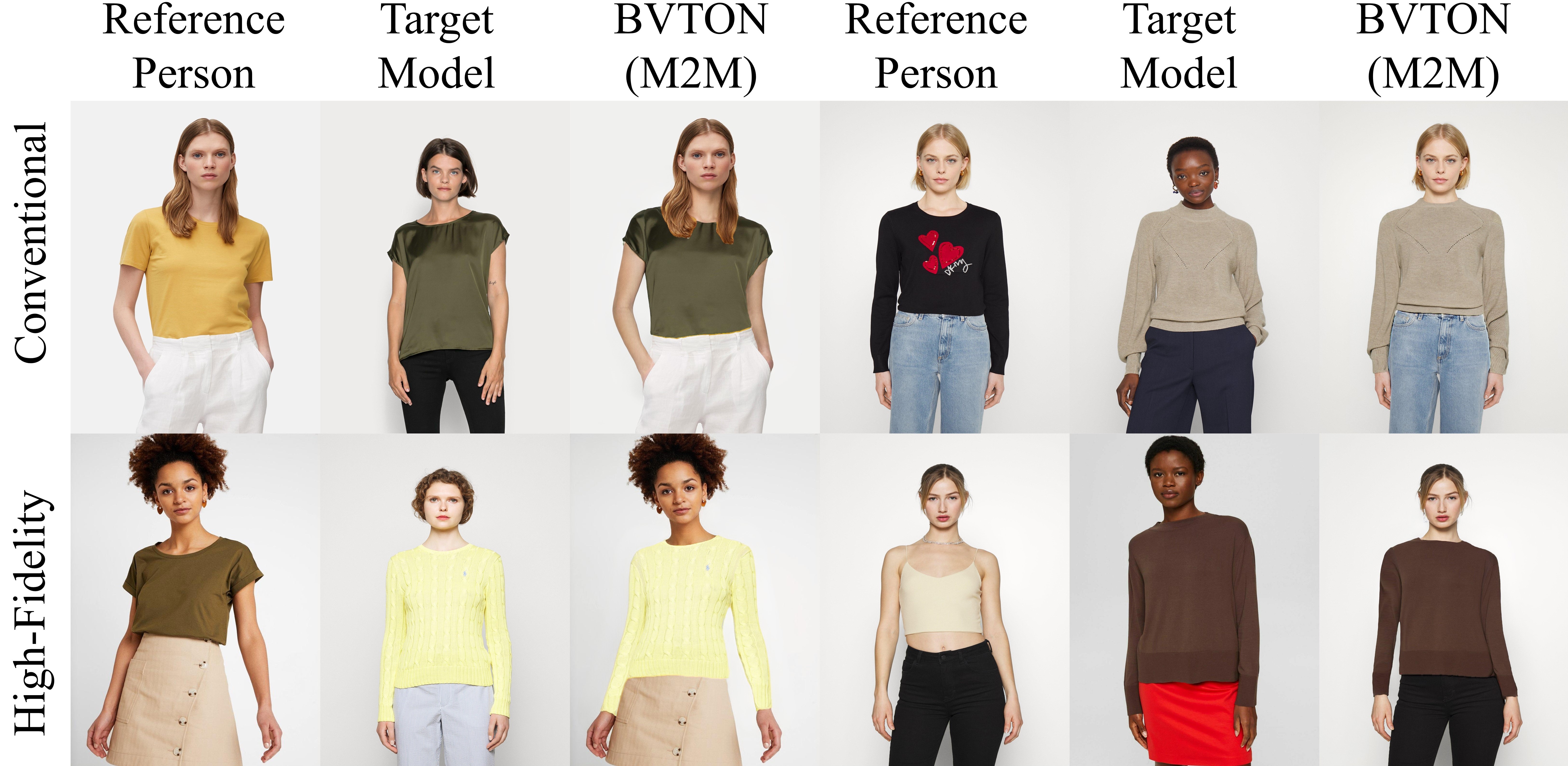}
\vspace{-10pt}
\end{center}
   \caption{Visualization of BVTON in model-to-model try-on application. \textbf{Figure in full resolution, please zoom-in for the details.}}
\label{fig:Model2model}
\vspace{-20pt}
\end{figure}

\subsection{Qualitative Results}
We conduct visual comparisons with three other different methods, which are the latest state-of-the-art works including the flow-based SF-VTON~\cite{SF-VTON}, the semantic-based methods RT-VTON~\cite{RT-VTON} and HR-VTON~\cite{HR-VTON}. Visual results on our in-domain test set, TEST1, are in Fig.~\ref{fig:Comparison}, and we also show the visual experiments on out-of-domain test sets, VITON and TEST2, in Fig.~\ref{fig:other_data}.

\noindent \textbf{Conventional Try-on Comparisons.}\quad
As given in Fig.~\ref{fig:Comparison} (row.~1, col.~1), all the methods generate quite reasonable red short-sleeve shirts on the reference person.  However, in terms of photo-realism and details, BVTON outperforms the baseline methods by a great margin. SF-VTON and HR-VTON generate weird arm shapes, which fail to decouple completely from the original clothing shape on the reference person. RT-VTON generates coherent body shapes, but we can observe that the sleeve boundaries are not as sharp as the result of BVTON. And in col.~2, only BVTON preserves the shape of puff sleeves. BVTON generates results with high clothing fidelity and better skin texture with clear neck shadow and clavicle details, ascribing to the large-scale fashion images. 
% BVTON also carves the clothing boundaries, which makes the synthesized results seemingly more photo-realistic than the trivial pixel-level overlay.
Results on the VITON and TEST2 test sets are given in Fig~\ref{fig:other_data}. 
% BVTON especially excels at capturing the clothing shape details, as in row.~3; BVTON can preserve the shape relationship of the puff-sleeve clothes while other methods fail to capture such details.
% Boosted by the abundant unpaired data, our L-MGM can predict unprecedentedly accurate semantic layout, which is crucial for actual commercial usage. 
% The ablation study on the training data in Fig.~\ref{fig:ablation_seg} shows that our superiority on semantic layout prediction credits to the usage of unpaired data rather than the network structure.

\noindent \textbf{High-Fidelity Try-on Comparisons.}\quad As given in Fig.~\ref{fig:Comparison} (row.~2), we can obviously see that only BVTON demonstrates the clothing fidelity by retaining the accurate clothes length. RT-VTON performs better in keeping the clothes length, but fails to synthesize stable clothes bottoms due to the ambiguous training scheme of paired data; the semantic prediction module in RT-VTON can only predict an average clothes bottom shape without extra guidance on the wearing styles during training. Besides, we can also observe the remarkable clothing shape modeling of BVTON in row.~2. Boosted by large-scale unpaired learning, our method accurately generates the hood (col.~2) while all other methods fail to predict that part.

\noindent \textbf{Model-to-model Try-on.}\quad Our unified framework adapts to model-to-model try-on with simple modification by first generating the canonical proxy from the target model, and then performing the identical inference process to fit the on-model clothes at the reference person, as given in Fig.~\ref{fig:Model2model}.

% Our framework is the first image based virtual try-on which also seamlessly supports model-to-model try-on application, while producing equally coherent and realistic results as the original setting. 

\begin{table}
\renewcommand\tabcolsep{3.0pt} % 调整表格列间的长度
\small

\begin{center}
\caption{Quantitative results on three different test sets, TEST1, TEST2, and VITON, which are all in $1024\times 768$. ``conv" and ``high" denote conventional and high-fidelity, respectively. We show the FID~\cite{FID}, LPIPS~\cite{lpips}, and SSIM~\cite{ssim}. BVTON is given in three settings: ``paired" uses paired L-MGM intead of canonical proxy, ``small" uses data only from PAIRED dataset, and ``full" uses additional large-scale fashion images (50k) to boost performance. ``*" indicates methods which are only for reference, not the main baselines; discussions are provided in supp..}
\begin{tabular}{cccccc}
\specialrule{.15em}{.05em}{.05em} 
Dataset&Method&FID (conv)&FID (high)&LPIPS&SSIM \\
\specialrule{.1em}{.05em}{.05em} 
\multirow{8}{*}{TEST1} & PF-AFN* & 9.892 & 30.421 & 0.114 & 0.778 \\
&DAFlow*  & 12.11 & N/A & 0.149 & 0.707 \\
&SF-VTON &9.522 &30.542 &{0.109} &0.780 \\
&RT-VTON  &{9.057} &{23.419} &0.116 &{0.767} \\
&HR-VTON  &11.852 &32.281 &0.146 &0.770 \\
\cmidrule(r){2-6}
&Ours{\tiny (paired)}& \underline{7.903} & 23.315 & \underline{0.098} & 0.812 \\
&Ours{\tiny (small)}&9.132 &\underline{19.328}&0.103 &\textbf{0.815} \\
&Ours{\tiny (full)}~  &\textbf{7.681} &\textbf{16.034} &\textbf{0.096} &\underline{0.814}  \\
\hline
\hline
\multirow{ 8}{*}{VITON}& PF-AFN* & 12.942 & 33.462 & 0.142 & 0.790 \\
&DAFlow* & 18.609 & N/A & 0.197 &0.714 \\
&SF-VTON  &12.814 &32.387 &\textbf{0.137} &0.793 \\
&RT-VTON  &{12.154} &{27.002} &0.148 &0.794 \\
&HR-VTON  &15.637 &34.316 &0.146 &{0.788} \\
\cmidrule(r){2-6}
&Ours{\tiny (paired)} & \underline{10.694} &26.455 &0.140 &\underline{0.806} \\
&Ours{\tiny (small)}&10.997 &\underline{23.611}&0.145 &0.802 \\
&Ours{\tiny (full)}~ &\textbf{10.318} &\textbf{19.796} &\underline{0.140} &\textbf{0.806} \\
\hline
\hline
\multirow{ 8}{*}{TEST2}& PF-AFN* & 10.283 & 33.075 & 0.142 & 0.779 \\
&DAFlow* & 11.438 & N/A & 0.163 & 0.674 \\&SF-VTON  &9.947 &34.797 &{0.138} &0.781 \\
&RT-VTON  &10.095 &{28.428} &0.143 &{0.772} \\
&HR-VTON  &{9.709} &34.843 &0.144 &0.772 \\
\cmidrule(r){2-6}
&Ours{\tiny (paired)} & \underline{7.766} &27.954 &\underline{0.101} &0.828 \\
&Ours{\tiny (small)}&8.188 &\underline{26.234}&0.106 &\textbf{0.829} \\
&Ours{\tiny (full)}~  &\textbf{7.661} &\textbf{24.058} &\textbf{0.101} &\underline{0.828} \\
\specialrule{.15em}{.05em}{.05em} 

\end{tabular}

\label{tab:Quant}
\vspace{-20pt}
\end{center}
\end{table}

%% In case the reviewer argues we didn't compare with other person2person sotas.
% Notably, the setting of model-to-model only shows the adaptability of our unified framework, we do not claim the superiority over other state-of-the-art works on model-to-model try-on.

\subsection{Quantitative Results}
Quantitative results are hard to evaluate for try-on tasks without official ground-truth of the reference person wearing different target clothes. The Fréchet Inception Distance is the widely-used metric for evaluating the try-on performance. Reconstruction metrics,
as addressed in \cite{PF-AFN,RT-VTON}, are not so suitable for virtual try-on, so we only list them here for reference. We use Learned Perceptual Image Patch Similarity (LPIPS), and Structural Similarity (SSIM) for reconstruction metrics.

Our quantitative results on three test sets (TEST1, VITON, and TEST2) are given in Tab.~\ref{tab:Quant}. BVTON outperforms the latest baselines in almost all metrics by a remarkable margin, with at most 5.319 in FID (conventional), 16.247 in FID (high-fidelity), 0.05 in LPIPS, and 0.056 in SSIM. By utilizing the strong magic of vastly available fashion images, the gap is straight-forward. In case of suspicion on unfair comparison, we also present results without extra large-scale unpaired data, namely BVTON{\tiny (small)}, which uses identical training data as all the baselines. Credited to canonical proxy, even with much smaller data size, BVTON demonstrates superiority over the remaining baselines, including BVTON{\tiny (paired)}.

\begin{figure}[htb]
\begin{center}
\includegraphics[width=0.85\linewidth]{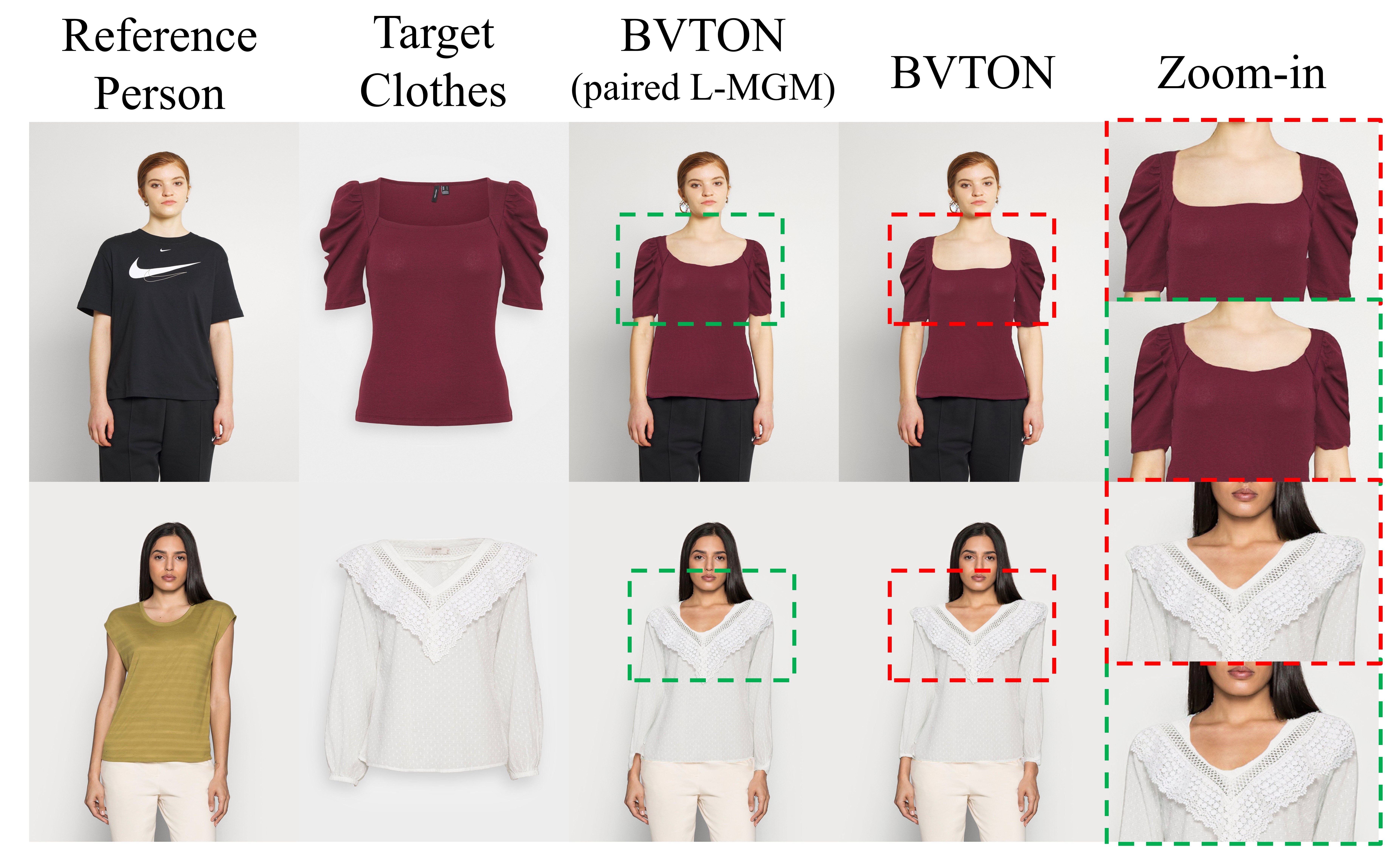}
\vspace{-10pt}
\end{center}
   \caption{Ablation study of canonical proxy with BVTON (L-MGM trained with paired data), namely BVTON {\tiny (paired)}, in the conventional setting. Quantitative results of BVTON {\tiny (paired)} is given in Tab.~\ref{tab:Quant}.}
\label{fig:ablation_seg}
\vspace{-10pt}
\end{figure}

\subsection{Ablation Study}
\begin{figure}[t]
\begin{center}
\includegraphics[width=0.85\linewidth]{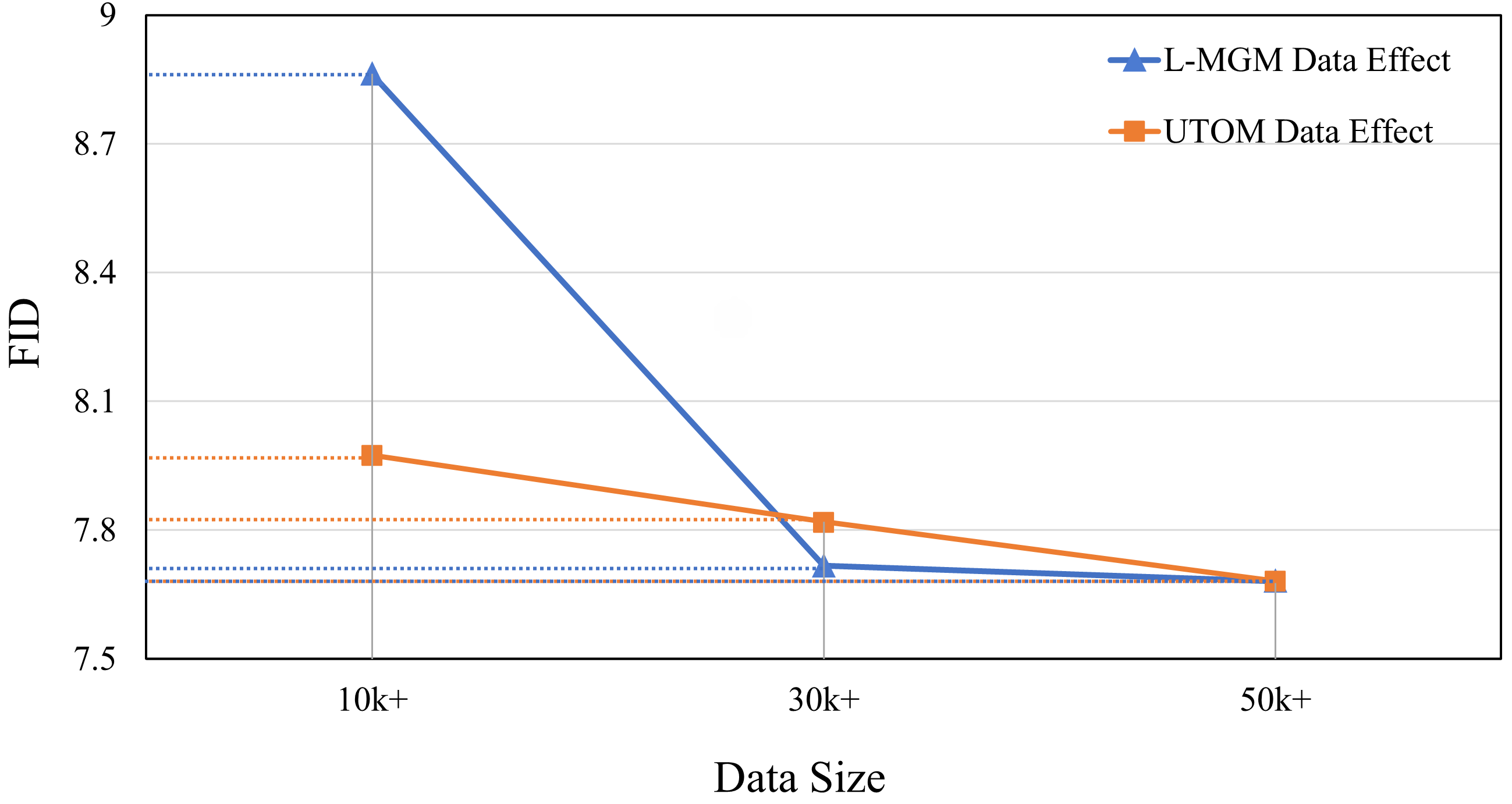}
\vspace{-5pt}
\end{center}
   \caption{Ablation study on the influence of unpaired data size on semantic prediction and image synthesis. Three data sizes are taken into consideration, which are 15,527, 32,971, and 50,415. FID scores decrease steadily along with the increaseing data size.}
\label{fig:data_influence}
\vspace{-10pt}
\end{figure}

 Our ablation study is conducted mainly to validate the effectiveness of our large-scale unpaired learning scheme. We first replace the L-MGM model with the one trained using only the  paired in-shop data to show the improvement in shape modeling with canonical proxy, as in Fig.~\ref{fig:ablation_seg} and Tab.~\ref{tab:Quant}. Then we show the gradual performance increase in FID along with the data size of unpaired data used in L-MGM and UTOM respectively, as in Fig.~\ref{fig:data_influence}.

\noindent \textbf{Effectiveness of Canonical Proxy.}\quad 
As given in Fig.~\ref{fig:ablation_seg}, BVTON (paired L-MGM) fails to predict the laces as well as the accurate collar shapes, while the BVTON full model accurately captures the non-standard shape details. Due to the ambiguity-free training scenario of our L-MGM, both the clothes bottoms and the clothing shape details such as laces and puff sleeves are well preserved through the semantic transformation. 
% In the paired setting, on the other hand, the aforementioned tiny clothes shape traits appear in different random styles, causing severe ambiguity in the L-MGM training.

%showing the superiority even with 1w data on tuck-out setting.

\noindent \textbf{Influence of Unpaired Data Size.}\quad 
In Fig.~\ref{fig:data_influence} we validate the relationship between the unpaired data size and the FID score. To validate the L-MGM model we will fix the UTOM model at 50,415 data, and vice versa.
We can see the FID score decreases drastically along with the data size for both L-MGM and UTOM, which demonstrates the strong motivation behind our large-scale unpaired learning.

%% file: sections/Conclusion.tex
\section{Conclusion}
We propose a principled framework, namely Boosted Virtual Try-on (BVTON), which leverages the large-scale unpaired learning to enhance the accuracy of semantic prediction and the quality of final try-on synthesis. Our framework transfers the on-model clothes into in-shop-like clothing shapes, constructing large-scale pseudo training pairs for the semantic learning.
Extensive results demonstrate the superior generalizability and scalability of BVTON over the state-of-the-art methods.

%% file: appendix.tex
\begin{figure}[htb]

\twocolumn[{
\renewcommand\twocolumn[1][]{#1}%

\begin{center}
  \centering
  \includegraphics[width=1\textwidth]{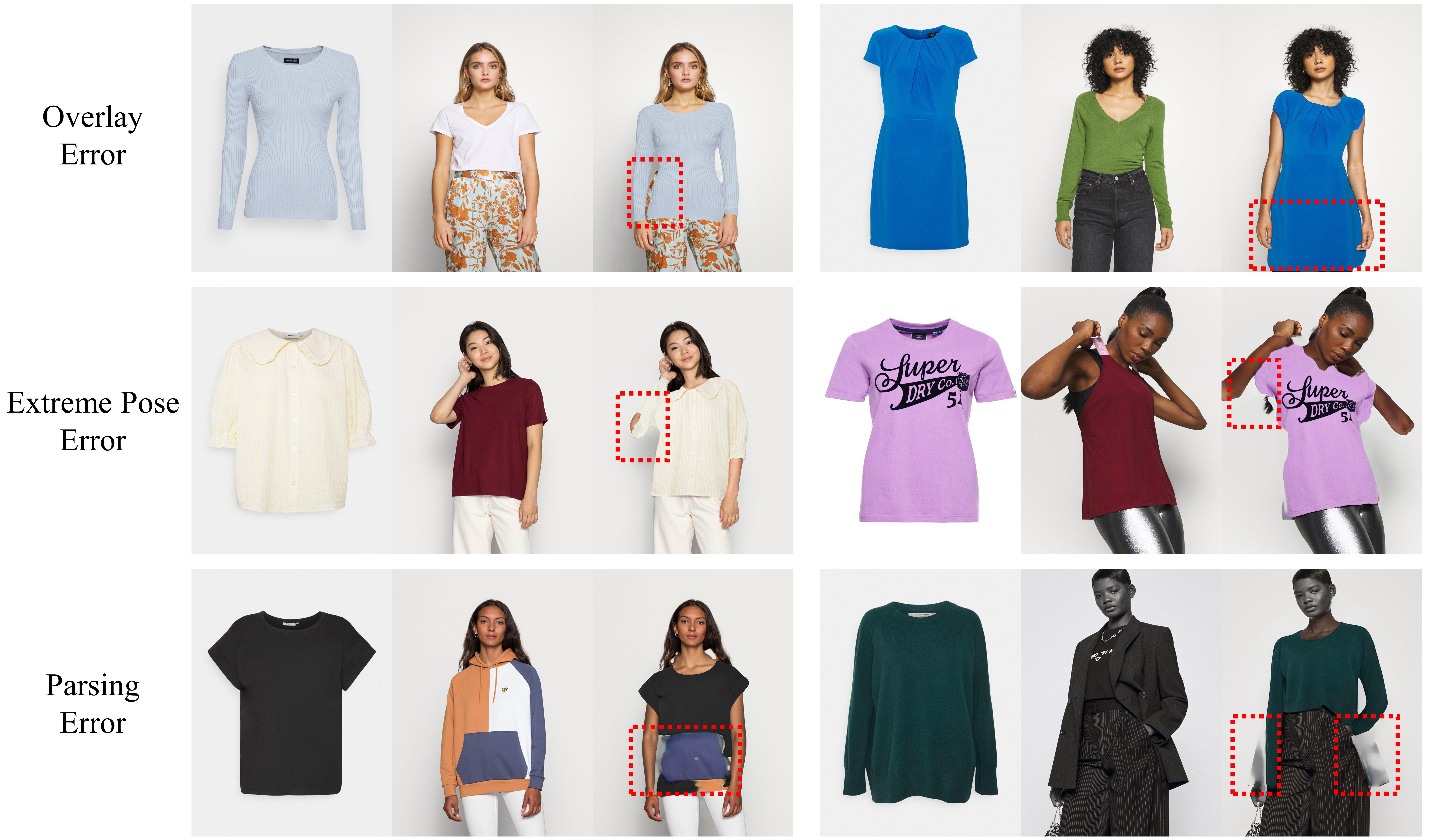}

\end{center}

\caption{Three failure cases including the extreme pose error, parsing error and the overlay error of the high-fidelity setting. }\label{fig:failure}

}]
\end{figure}

%%%%%%%%% ABSTRACT

\input{sections/Appendix}

%% file: sections/Appendix.tex
\section{Limitations}

In Fig~\ref{fig:failure} we show the failure cases of BVTON mainly in three aspects: \textbf{1)} Parsing error of the off-the-shelf parser in pre-processing, as in row.~3. On the left, the pocket of the reference person is mis-classified as retaining area; on the right, the back part of the jacket is not removed completely, causing dirty traces on the background. \textbf{2)} Extremely difficult pose of the reference person, as in row.~2. Folded arms are challenging for current 2D-based try-on methods to tackle. 3D-based methods with cloth simulation may better handle with the folded arms. \textbf{3)} The overlay error in high-fidelity try-on setting, as in row.~1. On the left, a small part of the bottom clothes is not completely occluded since the bottom clothing shape is agnostic to the network. A better agnostic representation can be proposed to fix this problem in future study. On the right, the target clothing image is overlong that causes the strange final try-on effect.

\section{Try-on Setting Validation}
\begin{figure}[htb]
\begin{center}
\includegraphics[width=1\linewidth]{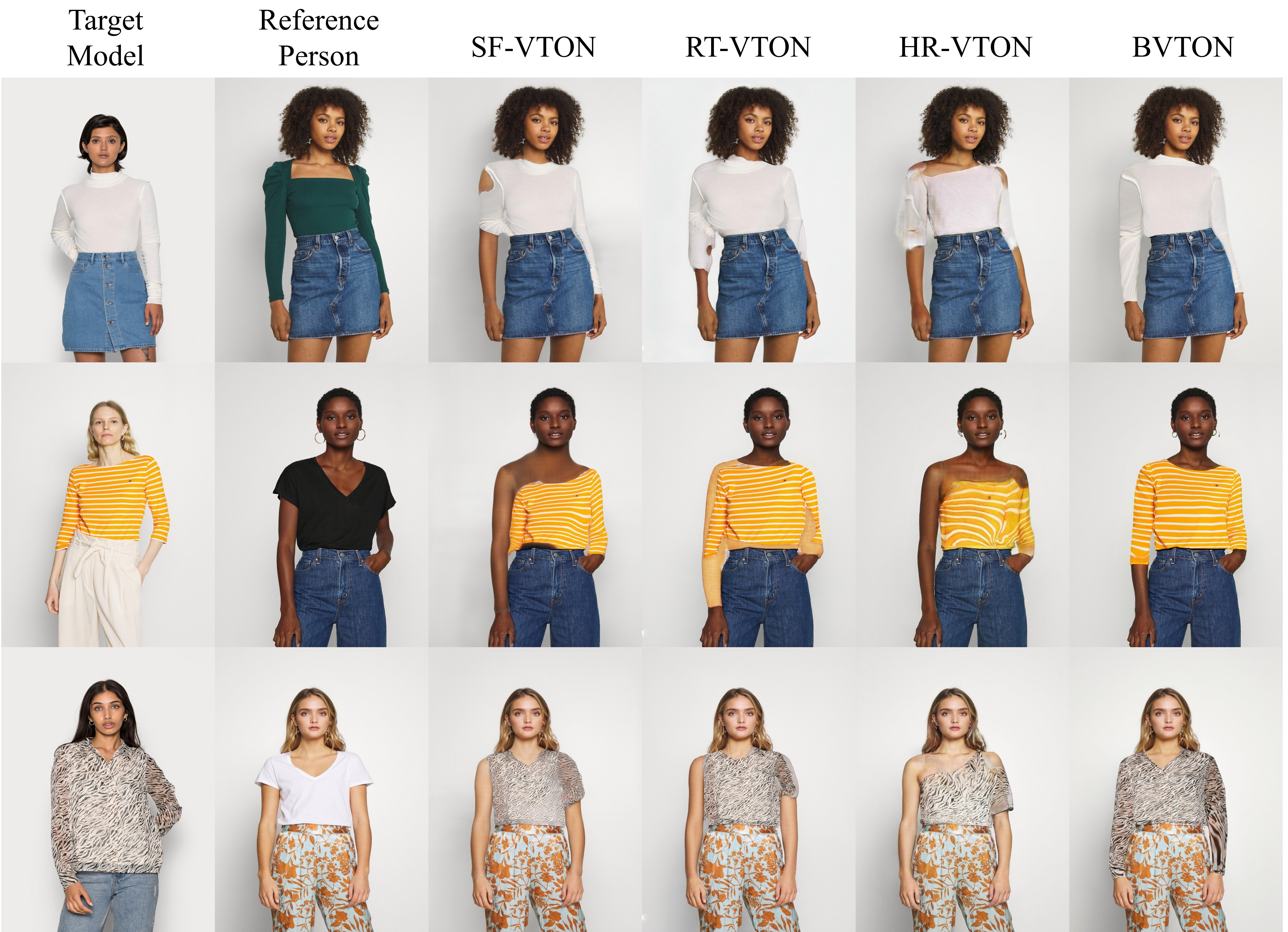}
\end{center}
   \caption{Baseline methods on model-to-model setting. Except BVTON, all the other cloth-to-model methods cannot perform model-to-model try-on by extracting the clothes from the model as target clothes.}
\label{fig:baseline_m2m}
\end{figure}

\begin{figure}[htb]
\begin{center}
\includegraphics[width=1\linewidth]{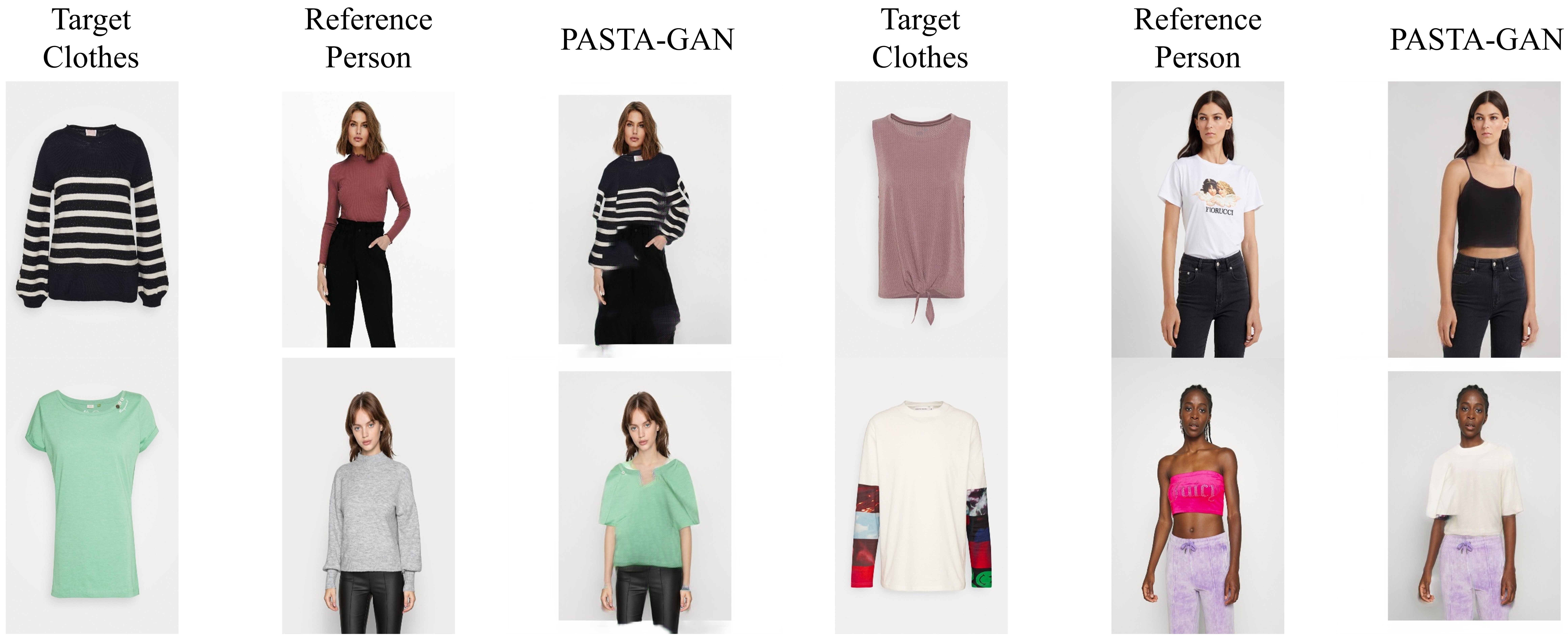}
\end{center}
   \caption{State-of-the-art method PASTA-GAN on cloth-to-model try-on setting. We apply the official release of PASTA-GAN and pretrained weights for demonstration.}
\label{fig:pasta-gan}
\end{figure}

Here we want to show that existing state-of-the-art cloth-to-model try-on works cannot perform model-to-model try-on by extracting the clothes from the model as the target clothes, and vice versa. We apply the pioneering cloth-to-model try-on works on model-to-model setting in Fig.~\ref{fig:baseline_m2m}, and the state-of-the-art model-to-model method, \ie, PASTA-GAN on cloth-to-model setting in Fig~\ref{fig:pasta-gan}.

\section{Loss Details}
{\it Segmentation loss} $\mathcal{L}_{CE}$. The pixel-wise cross-entropy loss which is computed between the ground-truth parsing obtained by off-the-shelf parser and the predicted logits.

{\it Conditional GAN loss} $\mathcal{L}_{cGAN}$. We use the multi-level patch-based discriminator to compute the cGAN loss, following pix2pixHD~\cite{pix2pixhd}. The squared distance is used to for the min max game.

{\it VGG loss} $\mathcal{L}_{vgg}$. To reconstruct with fine details and correct structure, vgg loss (also called perceptual loss) is always a good practice by computing the $L_1$ distance between the multi-level feature maps extracted by VGG network.

{\it Feature Matching loss} $\mathcal{L}_{feat}$. A classic design which works almost the same as vgg loss, which is computed with the $L_1$ distance between the multi-level feature maps extracted by the discriminator itself.
\section{Results on extra baselines}
\begin{figure}[htb]
\begin{center}
\includegraphics[width=1\linewidth]{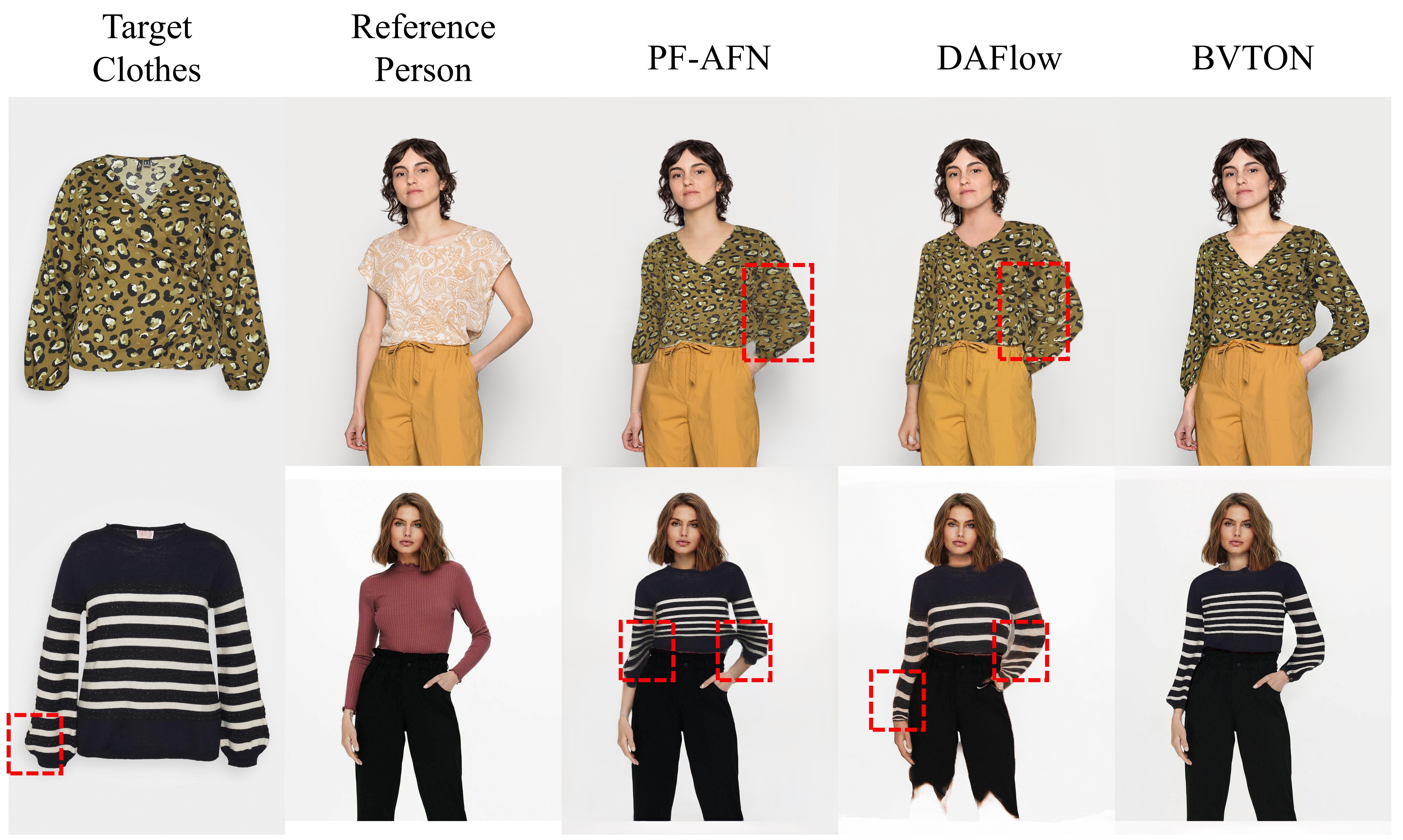}
\end{center}
   \caption{Extra qualitative results with DAFlow~\cite{DAFlow} and PF-AFN~\cite{PF-AFN}.}
\label{fig:extra}
\end{figure}

In case of the need for additional comparison, we here present the qualitative results of PF-AFN~\cite{PF-AFN} and DAFlow~\cite{DAFlow} in Fig.~\ref{fig:extra}.

\section{FAQ}
Here we summarize some possible questions for better understanding.

{\it BVTON uses 50k unpaired data. Is it unfair comparison?} The baseline methods are not capable of integrating unpaired data for training. Hence, if the baselines are trained with pseudo-paired data, then we change the original pipelines of their methods by plugging our CCM module into their frameworks. Hence, no unfair comparison is conducted. Additonally, we give results of BVTON{\tiny(small)} which uses identitcal training data as other methods while performing better.

{\it Why are PF-AFN and DAFlow not chosen as main baselines?} SF-VTON is the upgraded version of PF-AFN by only replacing the flow network structure and scoring higher, as shown in the paper of SF-VTON~\cite{SF-VTON}. Thus we choose SF-VTON as our main baseline method rather than PF-AFN. DAFlow also shows inferior results compared to SF-VTON in FID, as reported by their papers~\cite{SF-VTON,DAFlow}.

{\it Why is PASTA-GAN not compared?} PASTA-GAN can only perform model-to-model try-on as shown in Fig.~\ref{fig:pasta-gan}. BVTON mainly focuses on cloth-to-model try-on and demonstrates model-to-model try-on as an extra application, since the other baseline methods fail to perform model-to-model try-on as in Fig.~\ref{fig:failure}.

{\it 50k is too small to be called ``large-scale".} In virtual try-on scenario, 10k is the common data size~\cite{VITON_HD,VITON} so 50k is already very beneficial. BVTON is not restricted to only 50k data, more data can also increase performance.

{\it Is BVTON the first model-to-model try-on method?} The answer is No. BVTON is a cloth-to-model try-on method which also adapts to model-to-model try-on without retraining.

\section{Extensive Results}
Our extensive results are given in two folds: \textbf{1)} We conduct \textbf{frame-by-frame} try-on on out-of-domain video data (downloaded from the Internet) with target clothes from TEST2 test set. The videos are presented in \url{https://github.com/annnonymousss/BVTON}. \textbf{Notably, NO TEMPORAL OPERATIONS are used.} BVTON demonstrates convincing results even without temporal smoothing.
\textbf{2)} Extensive results on TEST1 are given in the following figures. We mainly present the image-based try-on results and the model-to-model try-on application. Single image samples are also shown for high-resolution viewing.

\begin{figure*}[t]
\begin{center}
\includegraphics[width=1\linewidth]{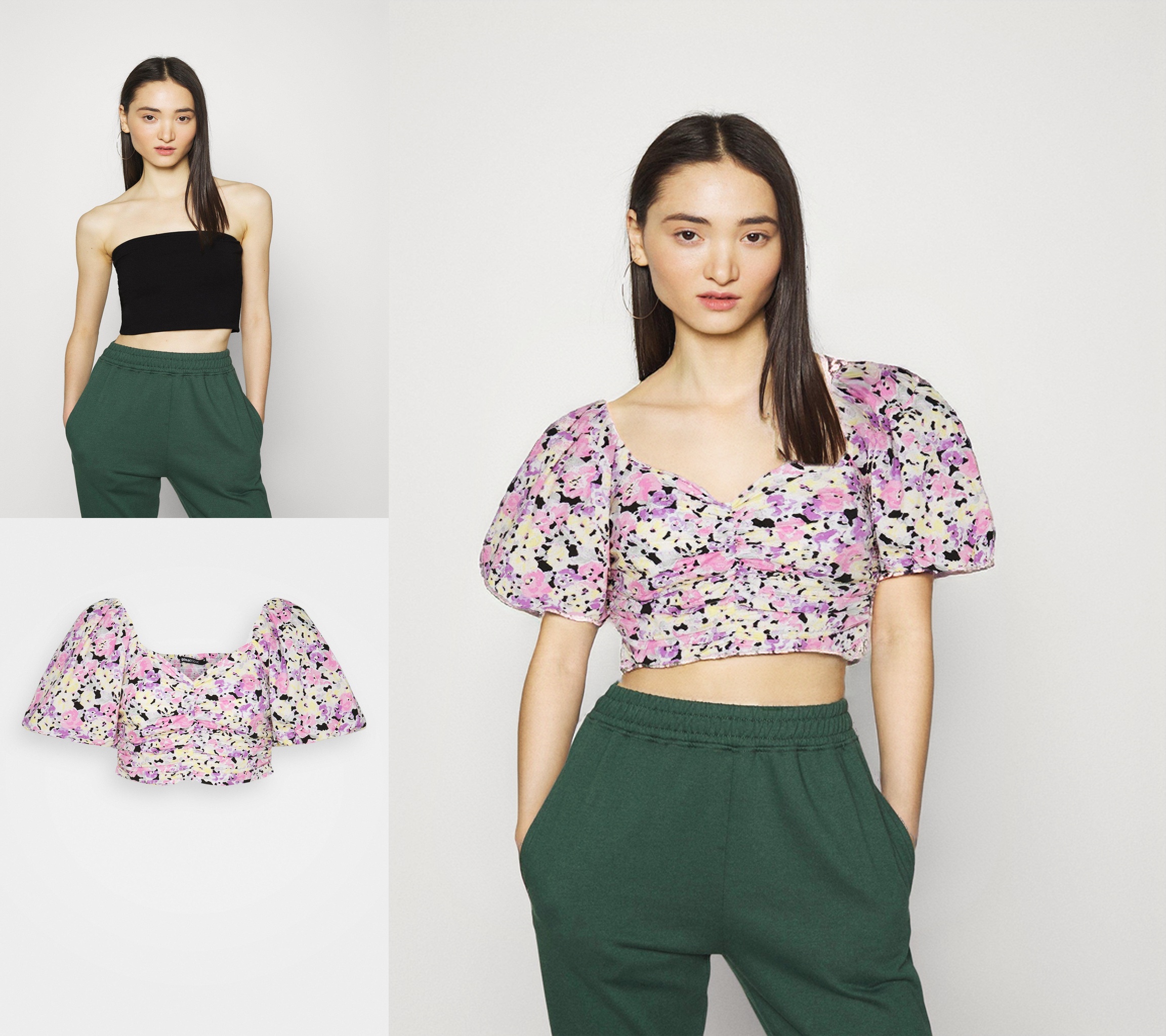}
\end{center}
   \caption{High-resolution sample of BVTON in conventional setting. Synthetic image is on the right with target clothes and reference person on the left.
   }
\label{fig:conven_single1}
\end{figure*}

% \begin{figure*}[t]
% \begin{center}
% %\fbox{\rule{0pt}{2in} \rule{0.9\linewidth}{0pt}}
% \includegraphics[width=1\linewidth]{Images/conven_single2.jpg}
% \end{center}
%    \caption{High-resolution sample of BVTON in conventional setting. Synthetic image is on the right with target clothes and reference person on the left.
%    }
% \label{fig:conven_single2}
% \end{figure*}

\begin{figure*}[t]
\begin{center}
\includegraphics[width=1\linewidth]{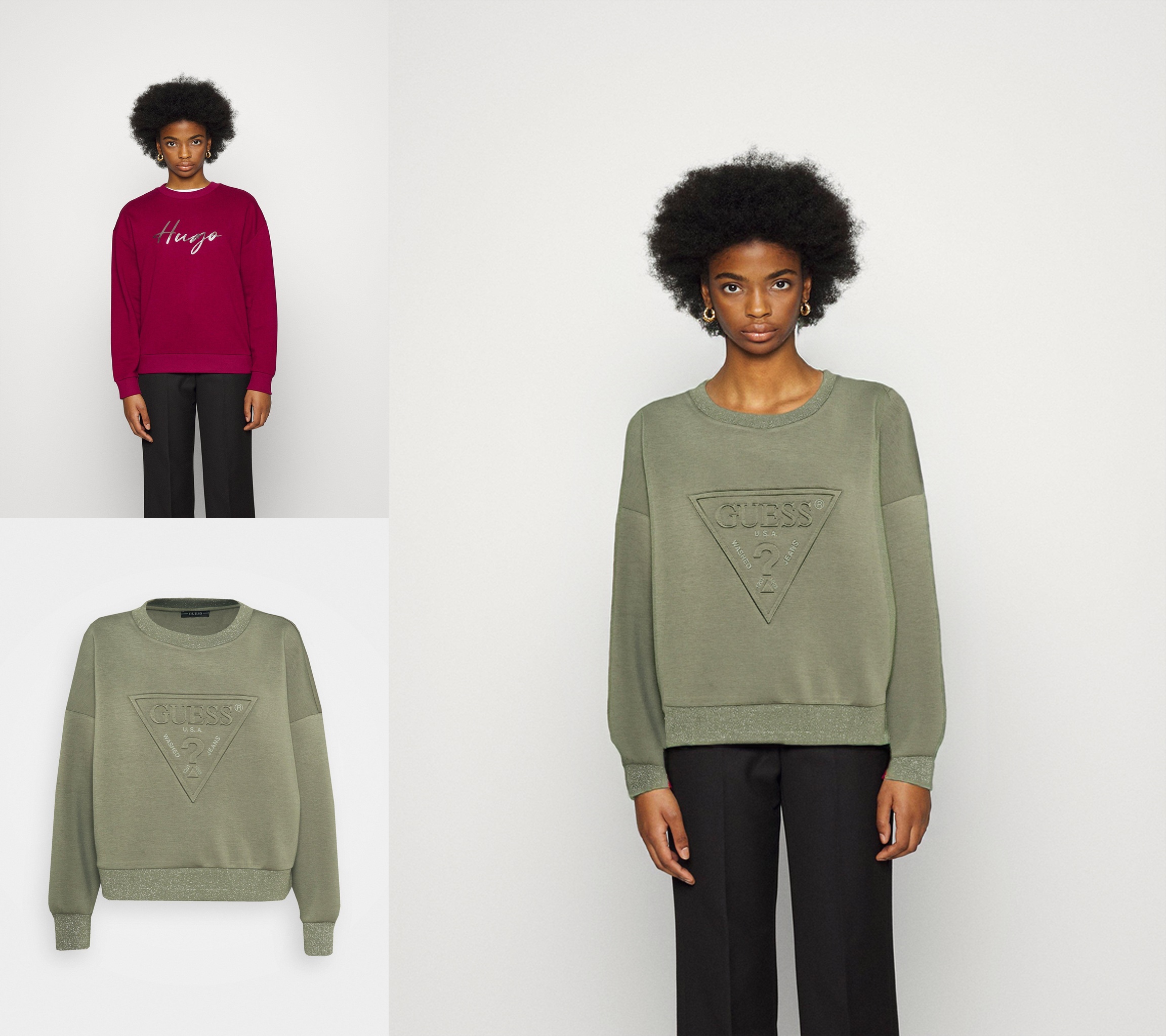}
\end{center}
   \caption{High-resolution sample of BVTON in conventional setting. Synthetic image is on the right with target clothes and reference person on the left.
   }
\label{fig:conven_single3}
\end{figure*}

\begin{figure*}[t]
\begin{center}
\includegraphics[width=1\linewidth]{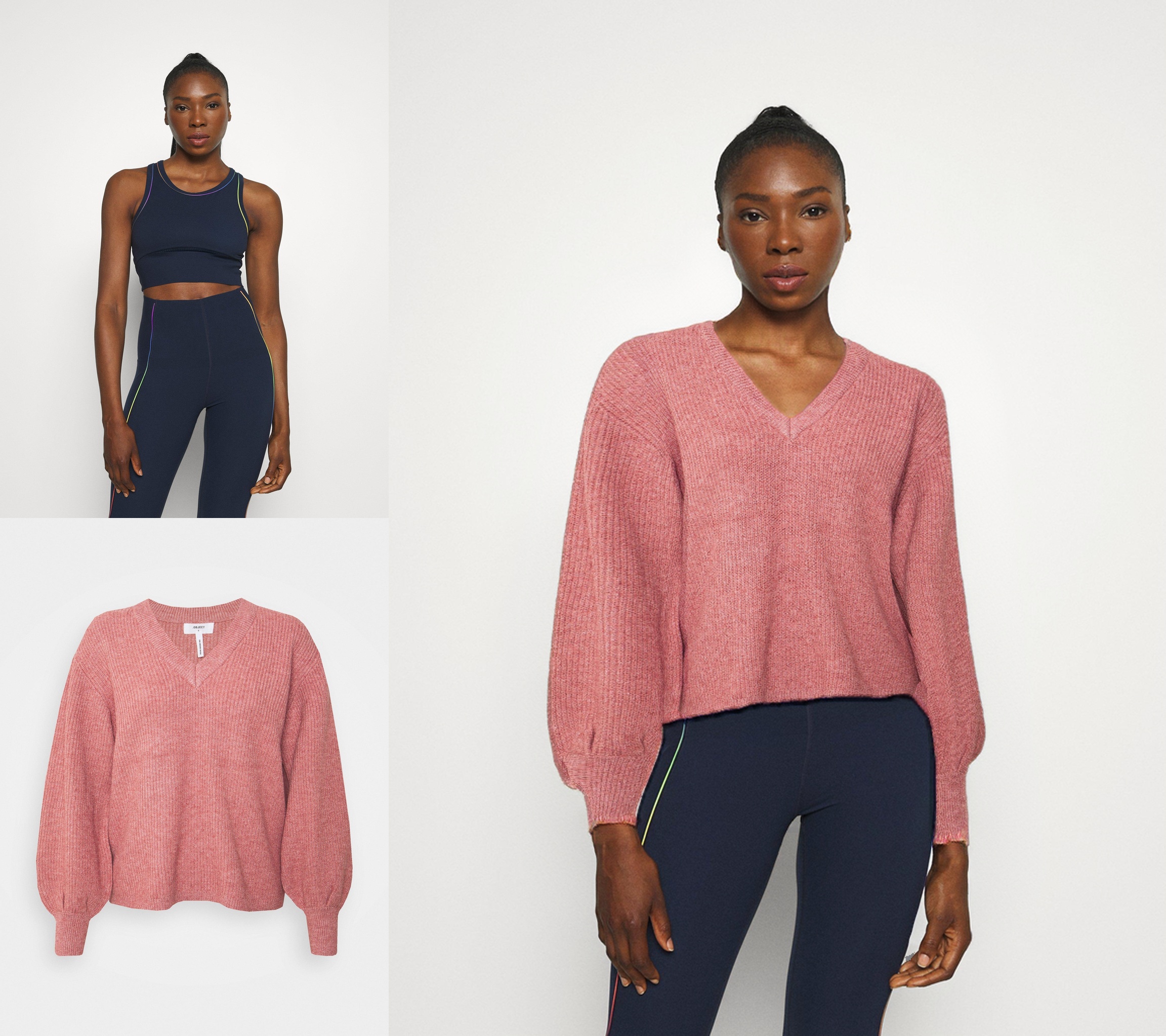}
\end{center}
   \caption{High-resolution sample of BVTON in high-fidelity setting. Synthetic image is on the right with target clothes and reference person on the left.
   }
\label{fig:HF_single1}
\end{figure*}

\begin{figure*}[t]
\begin{center}
\includegraphics[width=1\linewidth]{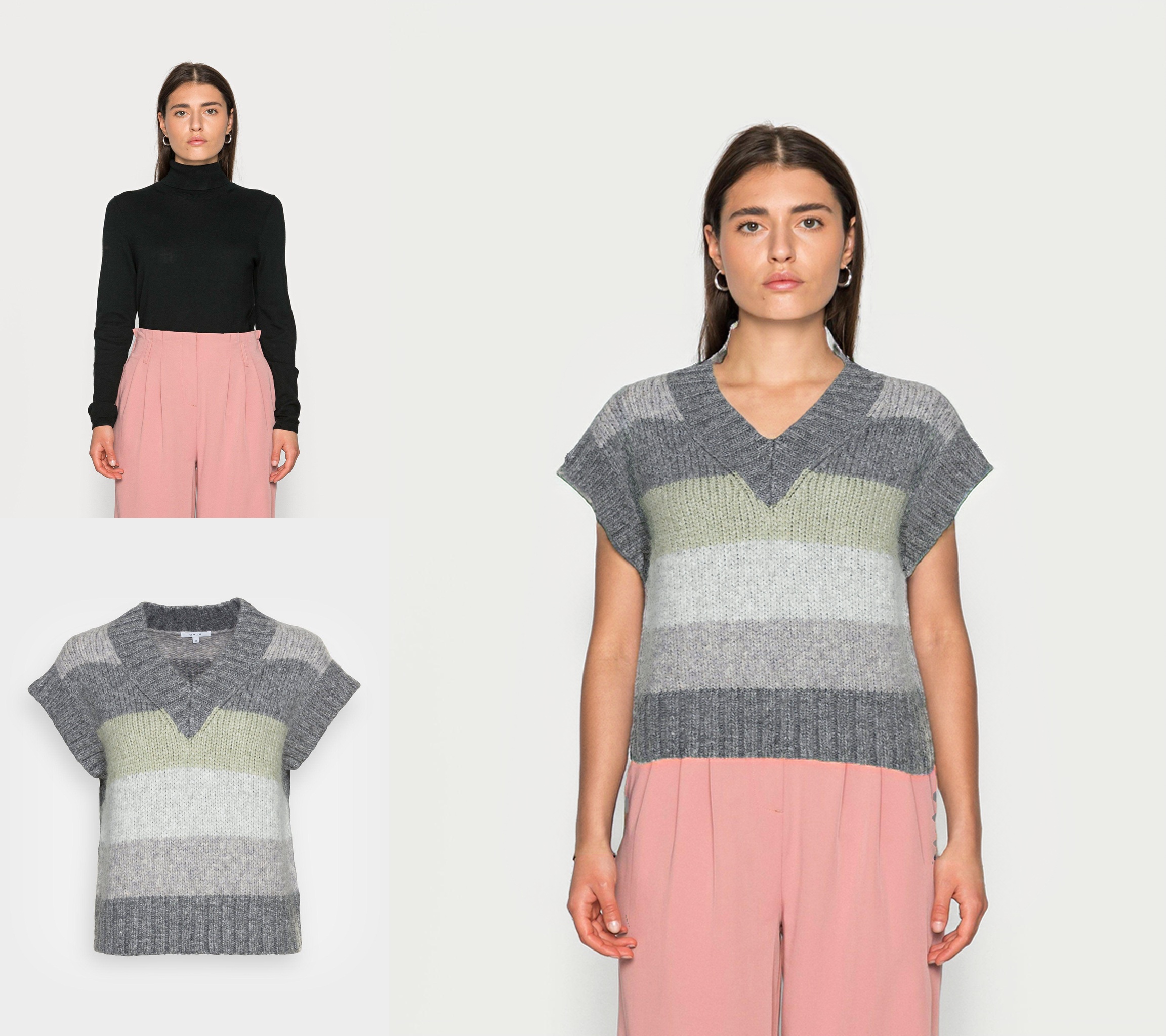}
\end{center}
   \caption{High-resolution sample of BVTON in high-fidelity setting. Synthetic image is on the right with target clothes and reference person on the left.
   }
\label{fig:HG_single2}
\end{figure*}

\begin{figure*}[t]
\begin{center}
\includegraphics[width=1\linewidth]{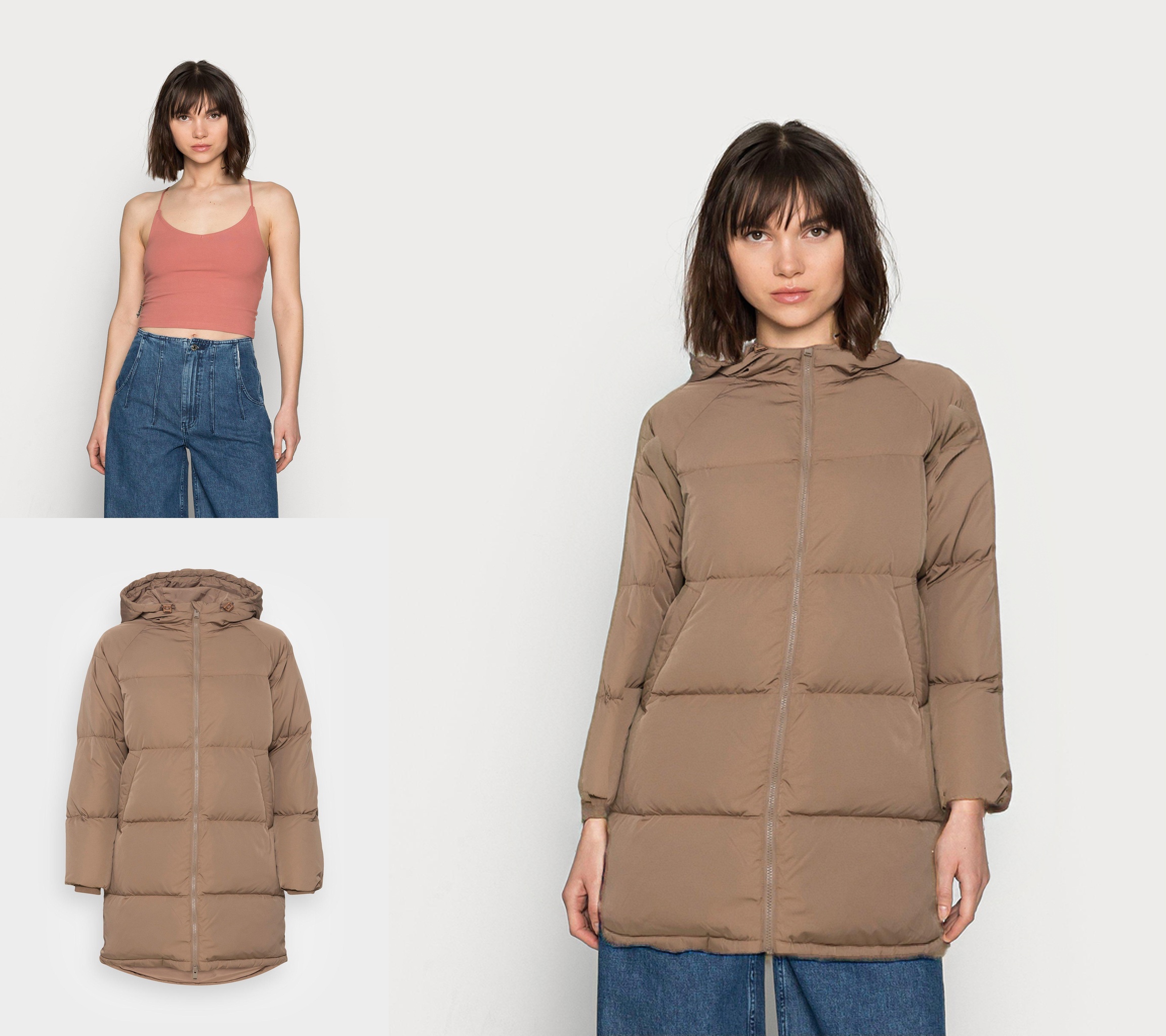}
\end{center}
   \caption{High-resolution sample of BVTON in high-fidelity setting. Synthetic image is on the right with target clothes and reference person on the left.
   }
\label{fig:HF_single3}
\end{figure*}

\begin{figure*}[t]
\begin{center}
\includegraphics[width=1\linewidth]{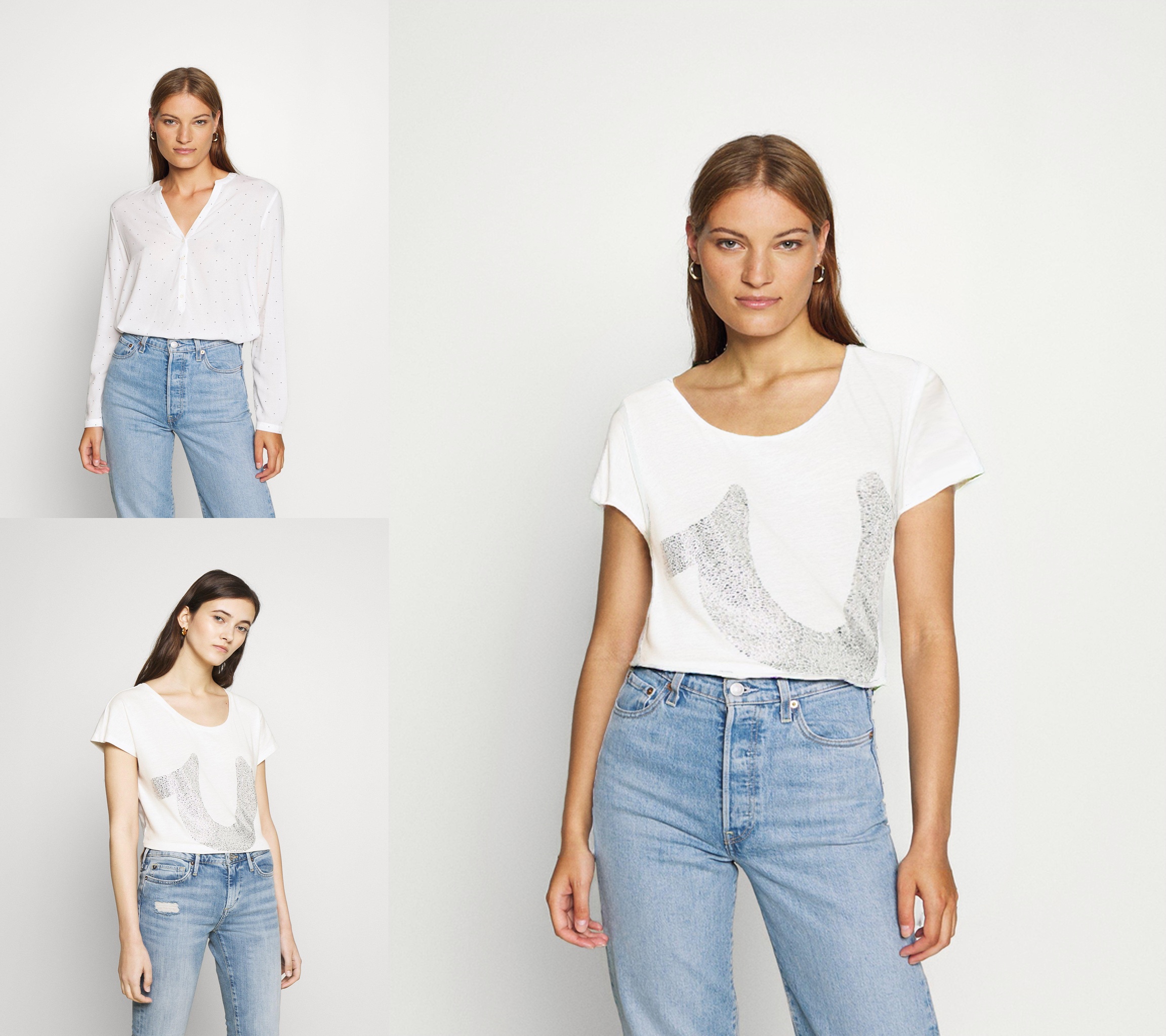}
\end{center}
   \caption{High-resolution model-to-model sample of BVTON in conventional setting. Synthetic image is on the right with target model and reference person on the left.
   }
\label{fig:M2M_in_single}
\end{figure*}

\begin{figure*}[t]
\begin{center}
\includegraphics[width=1\linewidth]{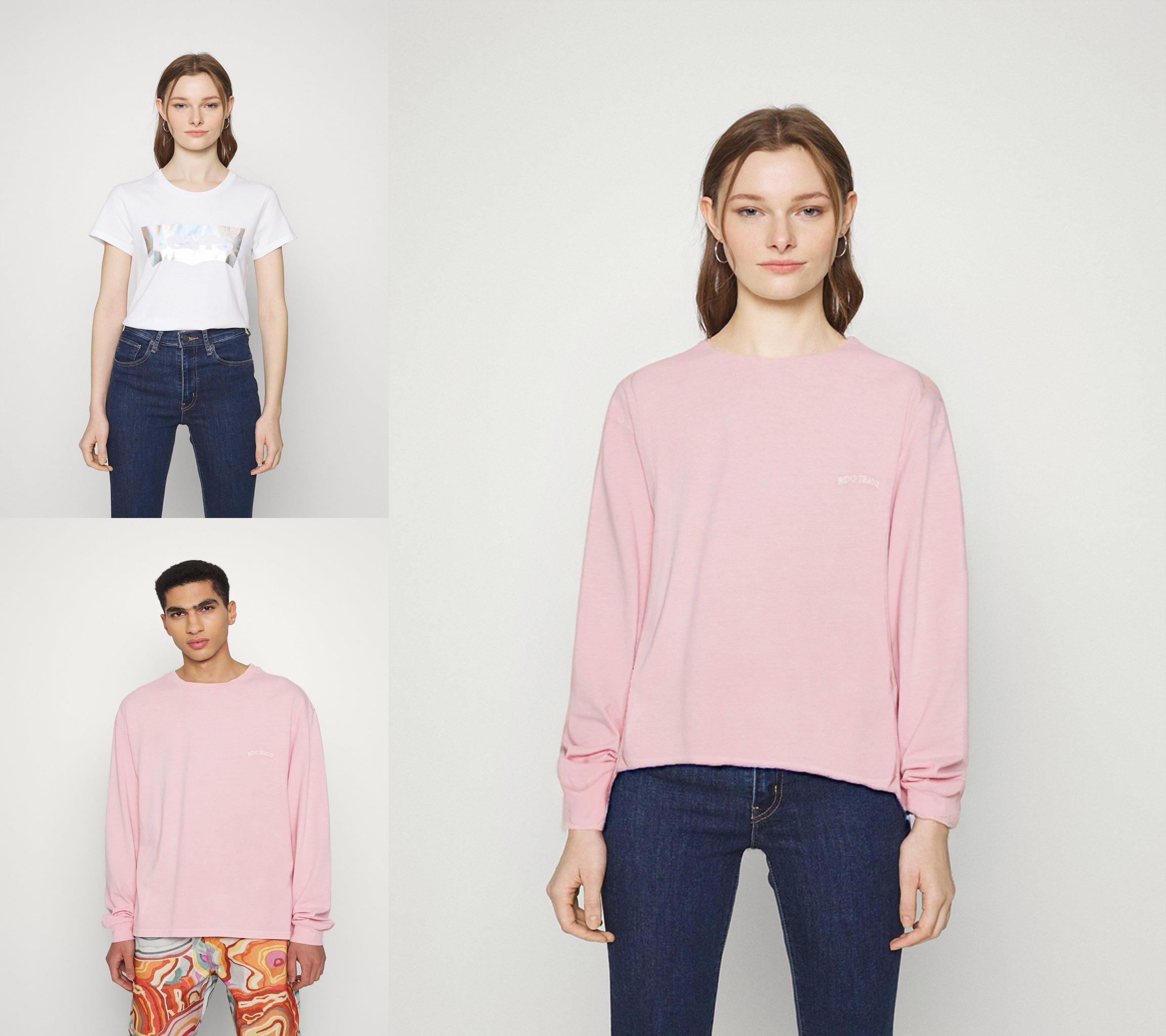}
\end{center}
   \caption{High-resolution model-to-model sample of BVTON in high-fidelity setting. Synthetic image is on the right with target model and reference person on the left.
   }
\label{fig:M2M_out_single}
\end{figure*}